\documentclass{article}

\usepackage[preprint]{neurips_2025}

\usepackage[utf8]{inputenc} 
\usepackage[T1]{fontenc}    
\usepackage{hyperref}       
\usepackage{url}            
\usepackage{booktabs}       
\usepackage[table]{xcolor} 
\usepackage{amsfonts}       
\usepackage{makecell}      
\usepackage{multirow}
\usepackage{adjustbox}
\usepackage{nicefrac}       
\usepackage{microtype}      
\usepackage{xcolor}         
\usepackage{xspace}
\usepackage{amsmath}
\usepackage{graphicx}
\usepackage{algorithm}
\usepackage{algpseudocode}
\usepackage{listings}

\definecolor{mygray}{gray}{0.85}
\definecolor{darkgreen}{RGB}{0, 150, 0}

\title{VisionThink: Smart and Efficient \\Vision Language Model via Reinforcement Learning}

\author{
    \vspace{-25pt}\\
    \textbf{Senqiao Yang\thanks{Equal Contribution}$^{\phantom{*}\,1}$\quad Junyi Li$^{*\,2}$\quad Xin Lai$^{*\,1}$ \quad Bei Yu$^1$\quad  Hengshuang Zhao$^2$\quad  Jiaya Jia$^{1,3}$} \\[5pt]
    $^1$CUHK ~~\quad\quad $^2$HKU~~\quad\quad  $^3$HKUST ~~\quad\quad \\[5pt]
    Codes and models:~\, \url{https://github.com/dvlab-research/VisionThink}
    \vspace{-8pt} \\
}

\newcommand{\mypara}[1]{\smallskip\noindent\textbf{#1}}

\begin{document}
\def\methodNAME{{VisionThink}\xspace}

\maketitle
\begin{abstract}
Recent advancements in vision-language models (VLMs) have improved performance by increasing the number of visual tokens, which are often significantly longer than text tokens.
However, we observe that most real-world scenarios do not require such an extensive number of visual tokens.
While the performance drops significantly in a small subset of OCR-related tasks, models still perform accurately in most other general VQA tasks with only 1/4 resolution.
Therefore, we propose to dynamically process distinct samples with different resolutions, and present a new paradigm for visual token compression, namely, \methodNAME.
It starts with a downsampled image and smartly decides
whether it is sufficient for problem solving. Otherwise, the model could output a special token to request the higher-resolution image.
Compared to existing Efficient VLM methods that compress tokens using fixed pruning ratios or thresholds, \methodNAME autonomously decides whether to compress tokens case by case. As a result, it demonstrates strong fine-grained visual understanding capability on OCR-related tasks, and meanwhile saves substantial visual tokens on simpler tasks.
We adopt reinforcement learning and propose the LLM-as-Judge strategy to successfully apply RL to general VQA tasks. Moreover, we carefully design a reward function and penalty mechanism to achieve a stable and reasonable image resize call ratio.
Extensive experiments demonstrate the superiority, efficiency, and effectiveness of our method.

\end{abstract}
\section{Introduction}

Recently, Vision-Language Models (VLMs)~\cite{li2024mini, li2023blip,liu2023improvedllava,chen2023sharegpt4v,Qwen-VL} have achieved remarkable performance in general visual question answering (General VQA) and various real-world scenarios by projecting and adapting visual tokens into the LLM space~\cite{touvron2023llama,achiam2023gpt,zhu2023minigpt,bai2023qwen}.
However, as the performance of VLMs continues to improve, the consumption of visual tokens increases exponentially. For example, a 2048$\times$1024 photo taken with a smartphone required 576 visual tokens in LLaVA 1.5~\cite{liu2023improvedllava}, but now requires 2,678 visual tokens in Qwen2.5-VL~\cite{bai2025qwen2.5vl}. Therefore, it is imperative to avoid the excessive use of visual tokens.

Numerous works on visual token compression have been proposed~\cite{wen2024efficient, shi2023upop, he2024zipvl,jian2023expedited,chen2024image, zhang2024sparsevlm}. Most approaches prune or merge a fixed number of visual tokens using predetermined thresholds. However, redundancy levels vary across different questions and images, leading to a natural question: \textit{Should we really apply a uniform token compression ratio across all scenarios?}

\begin{figure}[t]
\centering
\includegraphics[width=1.0\textwidth]{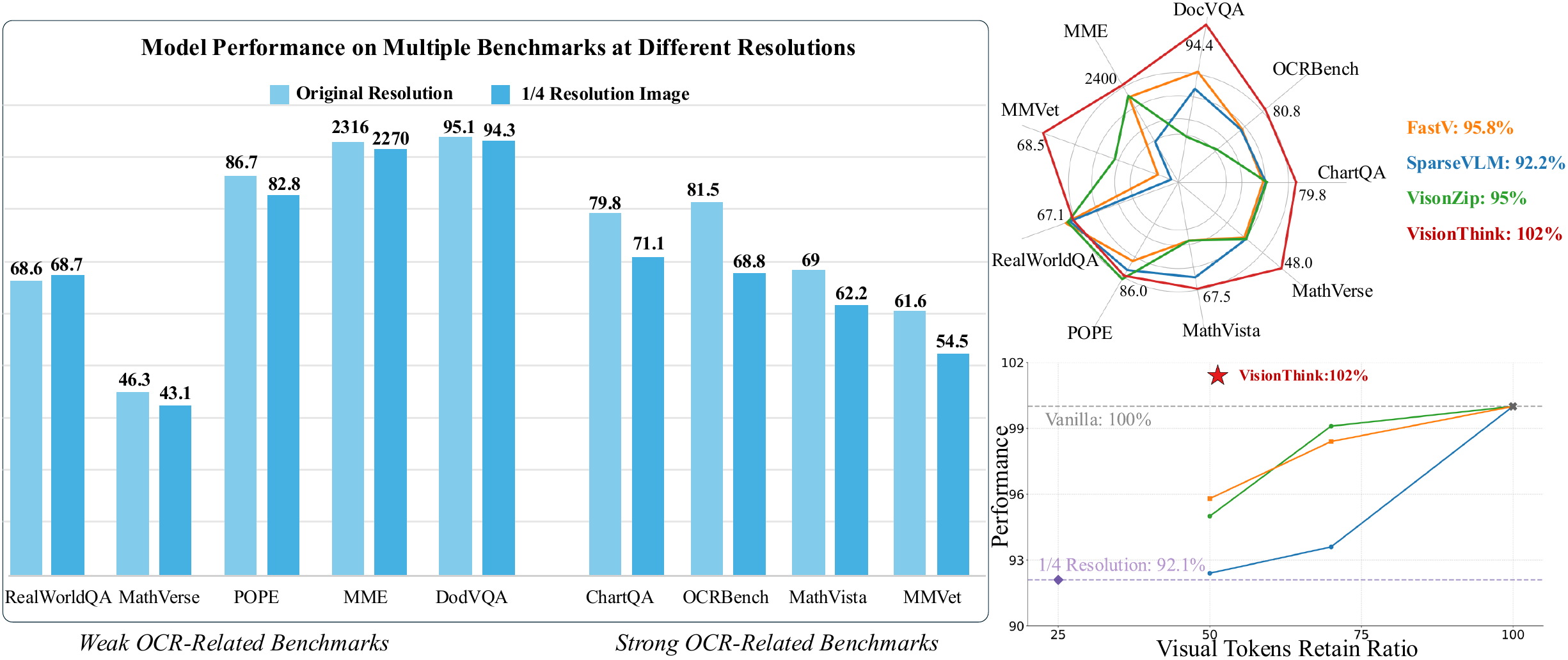}
\caption{\textbf{Our key observations and \methodNAME performance and efficiency}. \textbf{Left}: We find that in most general scenarios, even reducing visual tokens by a factor of four results in only minimal performance drop. However, token compression leads to a significant performance drop on strong OCR-related benchmarks. \textbf{Right}: Our VisionThink significantly outperforms previous work in both performance and efficiency. }
\label{figure:intro:benchmark}
\end{figure}

To answer this question, we simply reduced the image resolution to decrease the number of visual tokens and evaluated Qwen2.5-VL’s\cite{bai2025qwen2.5vl} performance on several benchmarks. As shown in the left of Fig.~\ref{figure:intro:benchmark}, we found that for most real-world scenarios, such as MME and RealWorldQA, even reducing the image resolution by a factor of four, which significantly cuts visual tokens by 75\%, has minimal impact on the model's performance.
However, as shown in the right of Fig.~\ref{figure:intro:benchmark}, for benchmarks such as ChartQA and OCRBench, which require detailed understanding and OCR-related capabilities, reducing the number of visual tokens leads to a significant drop in performance. 
Based on these observations, we find that most real-world questions do not require high-resolution images with long visual tokens, while a small subset of OCR-related tasks demand such detailed input much.
Therefore, there is significant potential for efficiency optimization if we can dynamically distinguish between samples that require high-resolution processing and those that do not.

In this paper, we propose~\methodNAME, a new EfficientVLM paradigm that leverages the model’s reasoning capabilities. Unlike prior methods that process full images and later discard redundant tokens,~\methodNAME directly inputs compressed visual tokens and allows the model to request the original high-resolution image when needed. This enables more efficient inference in most real-world scenarios, and meanwhile preserving performance on OCR-related tasks.

Although VisionThink offers a promising way to handle samples with varying levels of visual redundancy smartly, it still faces two key challenges:

\mypara{Effective Reinforcement Learning for General VQA.}
Conventional rule-based reinforcement learning algorithms, typically used to optimize reasoning process, struggle with the diversity and complexity of general VQA. To overcome this issue, we propose the LLM-as-Judge approach, enabling semantic matching. Experiments show performance improvement across several general VQA benchmarks, highlighting the potential to extend vision-based reinforcement learning beyond visual math reasoning to broader VQA tasks.

\mypara{Determine When High Resolution is Worth.}
To improve efficiency without compromising performance, the model must accurately determine when high-resolution input is necessary. We achieve this by carefully designing a balanced reward function to prevent the model from collapsing into always requiring high-resolution images or always using low-resolution images.
With this mechanism, VisionThink maintains strong performance on OCR benchmarks while delivering significant speed-ups on non-OCR benchmarks, achieving up to 100\% for DocVQA.

Overall, we present a simple yet effective pipeline—VisionThink. It introduces a new approach to visual token compression by dynamically determining compression based on the content of each sample, thereby achieving efficiency gains at the sample level. Consequently, it is compatible with other advanced spatial-level methods. We hope our work sheds new light on this area.

\section{Preliminary}
\subsection{Large Language Models and Reinforcement Learning}
Recent progress in improving the reasoning ability of large language models (LLMs)\cite{guo2025deepseekr1, jaech2024openai} has shown that Reinforcement Learning (RL) is an effective training approach. In this work, we use Group Relative Policy Optimization (GRPO)\cite{shao2024deepseekmath} as our training method. GRPO removes the need for a separate critic model by using group scores to estimate baselines. This reduces computation cost, improves training stability, and leads to faster and more reliable performance gains.

During the training process, GRPO samples a group of outputs $\{o_1, o_2, \cdots, o_G\}$ based on the given question $q$ from the old policy $\pi_{\theta_{old}}$ and then optimizes the policy model $\pi_{\theta}$ by maximizing the following objective:
\begin{equation}
\begin{split}
    \mathcal{J}_{GRPO}(\theta) & = \mathbb{E}_{[q \sim \mathcal{D}, \{o_i\}_{i=1}^G \sim \pi_{\theta_{old}}(\cdot|q)]}  \\
    \frac{1}{G}\sum_{i=1}^G & \left(\min\left(\frac{\pi_\theta(o_i |q)}{\pi_{\theta_{old}}(o_i |q)} A_i, \text{clip} \left( \frac{\pi_\theta(o_i |q)}{\pi_{\theta_{old}}(o_i |q)}, 1 - \epsilon, 1 + \epsilon \right)  A_i \right) - \beta \mathbb{D}_{KL}\left(\pi_{\theta} || \pi_{ref}\right)\right)
\end{split}
\label{eq:GRPO}
\end{equation}
\begin{equation}
    \mathbb{D}_{KL}\left(\pi_{\theta} || \pi_{ref}\right) = \frac{\pi_{ref}(o_i|q)}{\pi_{\theta}(o_i|q)}- \log\frac{\pi_{ref}(o_i|q)}{\pi_{\theta}(o_i|q)} - 1,
\end{equation}
where \( q \) represents the input questions drawn from the dataset \( \mathcal{D} \), and \( o \) denotes the generated text response. $\mathbb{D}_{\text{KL}}$ is the KL-divergence measure, while $\epsilon$ and $\beta$ are hyper-parameters. $A_i$ indicates the advantage, computed using a group of rewards $\{r_1, r_2, \ldots, r_G\}$ corresponding to the outputs within each group.

\subsection{Computation Complexity}
To evaluate the computational complexity of VLMs, we analyze key components, including the self-attention mechanism and the feedforward network (FFN). The total floating-point operations (FLOPs) are given by:
$$
\text{Total FLOPs} = T \times (4nd^2 + 2n^2d + 2ndm)
$$
where $T$ denotes the number of transformer layers, $n$ is the sequence length, $d$ is the size of the hidden dimension, and $m$ is the intermediate size of the FFN. 

This equation shows that computational complexity is heavily driven by the sequence length $n$. In general VLM tasks, the sequence length  $n$ is composed of $ n_{\text{sys}} + n_{\text{img}} + n_{\text{question}}$, where $n_{\text{img}}$ is often significantly larger than the other two components.
Hence, controlling the number of image tokens is key to achieving VLM efficiency.

\section{Methodology}
\subsection{Overview}
\begin{figure}[t]
\centering
\vspace{-0.3cm}
\includegraphics[width=1.\textwidth]{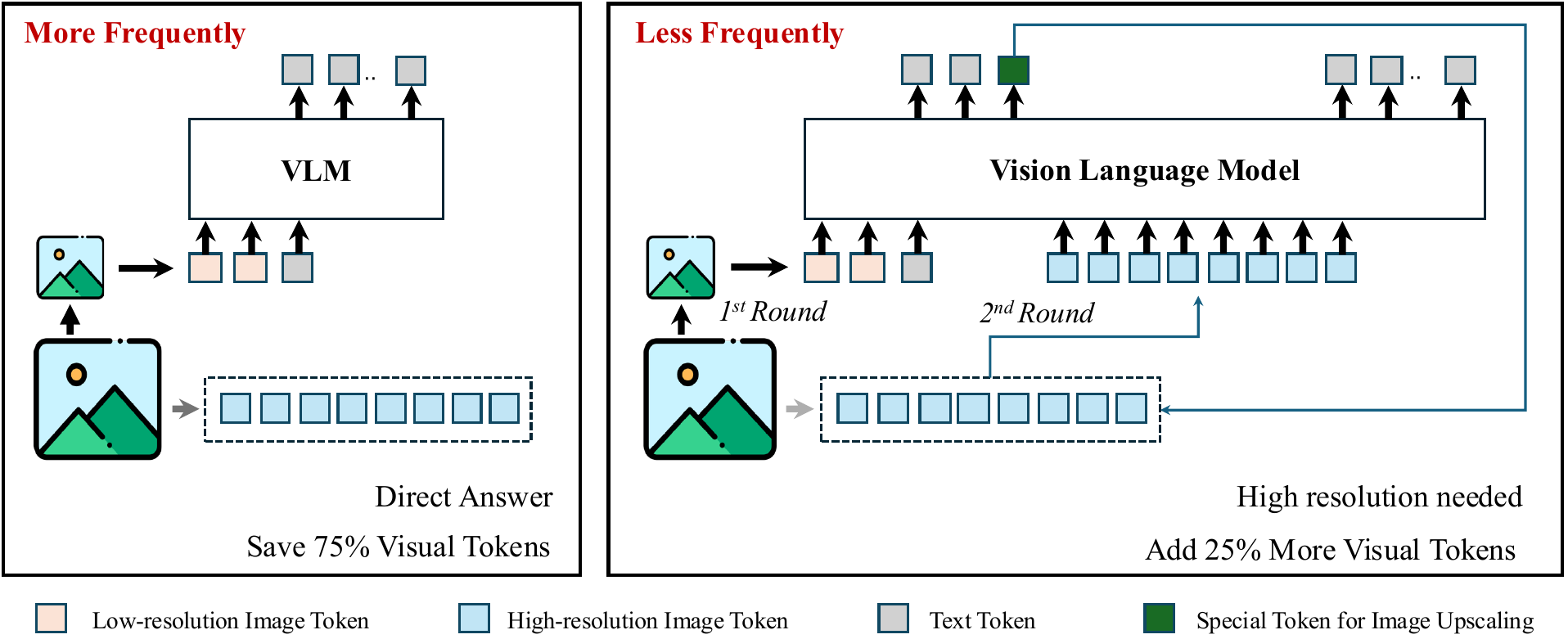}
\caption{\textbf{Framework of \methodNAME.} (a) The left image illustrates VisionThink processing an image with resolution reduced by a factor of four, where the VLM directly provides an answer. (b) The right image shows a case where the model detects insufficient information and requests a high-resolution image to answer the question.}

\vspace{-0.4cm}
\label{figure:framework}
\end{figure}

Our objective is to develop a smart and efficient VLM, capable of autonomously determining whether the information in the given image is sufficient to answer the question accurately.
 As shown in Fig.~\ref{figure:framework}, the pipeline first processes a low-resolution image to minimize the computataion cost. It then smartly requests original high-resolution inputs when the information in the downsampled image is insufficient to answer the question.
Ideally, this strategy maintains high performance while sharply reducing computational load.
To achieve this goal, we must address two key challenges:

\mypara{Effective RL on General VQA.}
Due to the diversity and complexity of general VQA, traditional rule-based RL algorithms are not directly applicable. To address this, we propose an LLM-as-Judge strategy, in which a large language model guides and evaluates the RL training process (Sec.~\ref{sec:Reward-general}). We further extend the Multi-Turn GRPO algorithm to suit our setting (Sec.~\ref{sec:multi-algorithm}).

\mypara{Enabling the model to decide when high resolution is necessary.}
The model must learn to assess whether a downsampled image contains sufficient information to answer the question if the original high-resolution image is required. So that the model could balance the efficiency and performance. To this end, we design a reward function that encourages optimal resolution decisions (Sec.~\ref{sec:reward}) and collect training data across multiple resolutions to support effective learning (Sec.~\ref{sec:data-prepare}).

\subsection{LLM-as-Judge for General VQA}
\label{sec:Reward-general}
\mypara{Challenges.} 
One of the central challenges in applying reinforcement learning to General VQA lies in evaluating model responses, especially when answers are open-ended or context-dependent. Most existing multi-modal RL efforts remain limited to structured tasks such as visual math, where ground-truth answers can be easily defined and verified via rules or exact matching. However, this approach breaks down in General VQA settings, where the diversity and ambiguity of valid answers make rule-based verification infeasible.

\mypara{Pure Text Accuracy Judgement.} 
To address this, we employ an external LLM as a judgment evaluator. Leveraging its broad knowledge and language understanding, the LLM assesses the correctness of model outputs in a human-aligned and flexible manner. 
Importantly, the evaluation is conducted purely in text by comparing the model’s answer with the ground-truth. This design avoids biases from visual content and the limitations of VLM performance. Furthermore, to minimize potential misjudgment by the evaluator, the reward is discrete (either 0 or 1) rather than continuous. The detailed judgment prompt is shown in Appendix~\ref{sec:supp-prompt-details}.

\mypara{Effectiveness. } 
The LLM-as-Judge is flexible, one advantage is that most of the SFT data could be used.  To verify the effectiveness of our proposed LLM-as-Judge, we collected 130K samples, which can be directly used to train the model with GRPO, without requiring any cold-start process. 
The results show significant improvement compared to the base model, Qwen2.5VL-Instruct. Further details are provided in Appendix~\ref{sec:supp-scaleup}.

\subsection{Mutli-Turn Training Algorithm}
\label{sec:multi-algorithm}
\mypara{Multi-Turn GRPO. }
In our \methodNAME framework, we first input the question and the downsampled image into the VLM. If the information is insufficient to answer the current question, the model will autonomously request a higher-resolution image and generate a new response. This process is essentially a multi-turn interaction.
Therefore, we extend the original GRPO~(Eq.~\ref{eq:GRPO}) to a multi-turn GRPO, as shown in Eq.~\ref{eq:multi-turn-GRPO}:

\begin{equation}
\begin{split}
\mathcal{J}_{GRPO}(\theta) = \, &
\mathbb{E}_{q \sim \mathcal{D}, \{ o_i \}_{i=1}^{G} \sim \pi_{\text{old}}( \cdot| q; \mathcal{I})} \Bigg[\frac{1}{G} \sum_{i=1}^{G} \frac{1}{\sum_{t=1}^{|o_i|} \mathbb{I}(o_{i,t})} \sum_{t=1}^{|o_i|}\mathbb{I}(o_{i,t}) \\ 
\cdot \min & \Bigg(p_{i,t}\hat{A}_{i,t}, \text{clip} \Bigg(p_{i,t}, 1 - \epsilon, 1 + \epsilon \Bigg) \hat{A}_{i,t} \Bigg) - \beta \mathbb{D}_{KL} \left[ \pi_{\theta} || \pi_{\text{ref}} \right]
\Bigg],
\end{split}
\label{eq:multi-turn-GRPO}
\end{equation}
where $p_{i,t}=\frac{\pi_{\theta}(o_{i,t} | q, o_{i,<t}; \mathcal{I})}{\pi_{\text{old}}(o_{i,t} | q, o_{i,<t}; \mathcal{I})}$, and $\mathbb{I}(o_t)$ is the token loss masking operation such that $\mathbb{I}(o_t)=1$ if $o_t$ is the generated token from LLM and $\mathbb{I}(o_t)=0$ if $o_t$ is the response token from the called tools.
Intuitively, we masked all text and image tokens from the user and performed optimization solely based on the multi-turn output tokens generated by the VLM.

\mypara{How does the model signal the need for a high-resolution image? }
To determine when the model requires a high-resolution image, we modify the prompt to instruct the model to output specific special tokens. Notably, this is a non-trivial process because our training does not introduce any cold-start phase, which leads to a performance drop in general VQA (Appendix~\ref{sec:supp-no-cold}).  Therefore, selecting an appropriate and effective prompt at the early stage of training is crucial.
The prompt must ensure that the model is capable of outputting the required special tokens during multi-turn rollouts in a zero-shot setting. Otherwise, GRPO will fail to optimize correctly due to the absence of gradients. We conduct comparative ablation studies in Appendix~\ref{sec:prompt-influence} and find that the Agent Prompt recommended by Qwen-2.5VL~\cite{bai2025qwen2.5vl} suits \methodNAME best. The prompt details are provided in Appendix~\ref{sec:supp-prompt-details}.

\subsection{Reward Design}
\label{sec:reward}
Different reward functions can lead the model toward different optimization directions and final performance outcomes. The reward function in our \methodNAME framework consists of three components:
\begin{equation}
\mathcal{R}_{\text{overall}} = \mathcal{R}_{\text{accuracy}} + \mathcal{R}_{\text{format}} - \mathcal{P}_{\text{control}},
\end{equation}
where $\mathcal{R}$ represents the reward and $\mathcal{P}$ represents the penalty.

\mypara{Accuracy Reward.}
The accuracy reward is designed following Sec.\ref{sec:Reward-general}, using LLM-as-Judge, where 0 indicates an incorrect answer and 1 indicates a correct answer.

\mypara{Format Reward.}
To maintain the model’s instruction-following capability and ensure that the trained model can more accurately call the image resize function, we apply a format reward. 
Specifically, we require the reasoning process to be enclosed in ``<think></think>'' tags, the final answer in ``<answer></answer>'' tags, and the function call to conform to the JSON format specified in Appendix B.
If any of these formats are incorrect, the format score is 0. Only when all formats are correct can the model achieve the full format score of 0.5.

\begin{figure}[t]
\centering
\includegraphics[width=\textwidth]{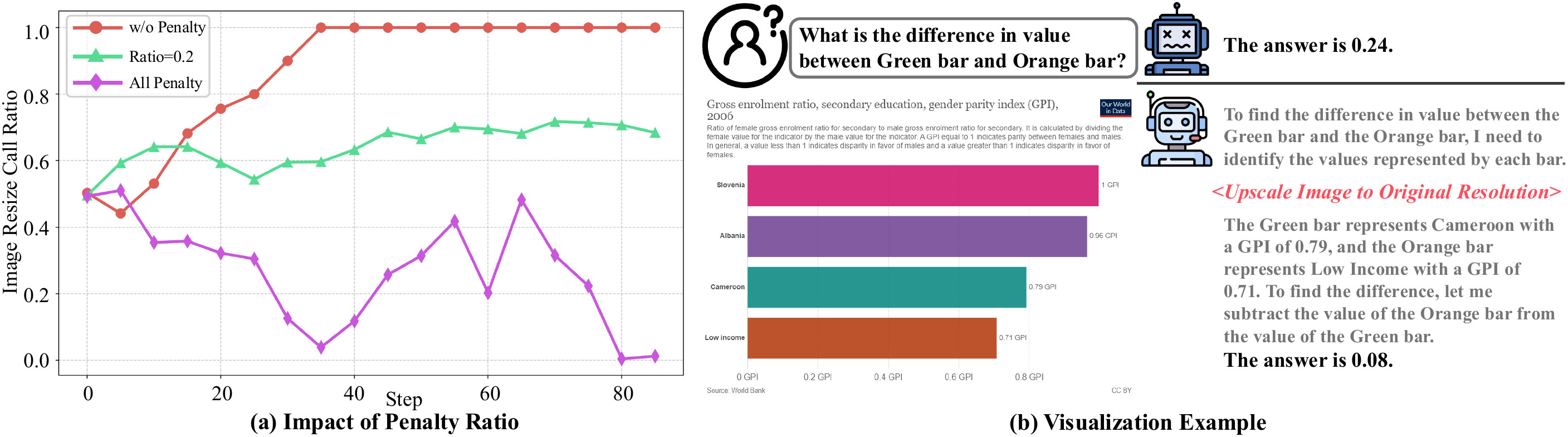}
\caption{(a) Impact of the Penalty Ratio. Applying a penalty to all resize image requests or removing the penalty entirely will both lead to model collapse.
 (b) VisionThink correctly solves OCR-related problems by autonomously requesting high-resolution images. }
\label{figure:analyze-penalty}
\vspace{-1cm}
\end{figure}

\mypara{Penalty Control.} The design of the penalty is a key component of the reward function.
As shown in Fig.~\ref{figure:analyze-penalty}(a), since using high-resolution images generally improves performance, without any penalty, the model tends to collapse into always requesting high-resolution images. To prevent this, we initially followed Search-R1~\cite{jin2025search} and applied a 0.1 penalty for correct answers that relied on high-resolution images.
However, this approach causes the model to favor direct answers, leading to a collapse where the model relies solely on direct answers, as indicated by the purple line in Fig.~\ref{figure:analyze-penalty}. The reason is that even blurry, low-resolution images sometimes allow the model to guess the correct answer, and the 0.1 penalty unintentionally reinforced this preference for direct answering.

To address this, we introduce a threshold to control the phenomenon of ``lucky guesses''. When the probability of correctly answering with a low-resolution image is low, we apply a 0.1 penalty to direct answers to encourage high-resolution requests; conversely, when the probability is high, we penalize high-resolution requests with a 0.1 penalty.
In summary, the  penalty is designed as below:
\begin{equation}
\mathcal{P}_{control}=0.1\cdot\left[\mathbf 1_{\text{direct}}\mathbb{I}(r<\theta)+\mathbf 1_{\text{high}}\mathbb{I}(r\ge\theta)\right],\qquad
r=\frac{C_{\text{direct}}}{C_{\text{direct}}+C_{\text{high}}},
\label{eq:penalty}
\end{equation}
where $C_{\text{direct}}$ and $C_{\text{high}}$ are the correct-answer counts for low- and high-resolution inputs, respectively, and $\mathbf 1_{\text{action}}$ is the indicator of the chosen action, and we set $\theta$ as 0.2 here. We will discuss the impact of the threshold in Appendix~\ref{sec:prompt-influence}.

\begin{table*}[t]
  \centering
  \caption{\textbf{Effective Performance Compared to the Sota Model.} Our model is based on Qwen2.5-VL-7B-Instruct. VisionThink$\ddag$ represents a model trained on general VQA tasks using full image resolution with the LLM-as-Judge strategy, which does not contain efficiency capabilities. Qwen2.5-VL-7B$^{*}$ reports the results evaluate by lmms-eval\cite{zhang2024lmms}.}
  \begin{adjustbox}{width=\textwidth}
  \renewcommand{\arraystretch}{1.5}
  \setlength{\tabcolsep}{2pt}
  \begin{tabular}{@{}l|ccccccccc@{}}
    \toprule
    \multirow{2}{*}{Method} & \textbf{MMMU} & \textbf{MMMU-Pro} & \textbf{MMBench} &\textbf{RealWorldQA} & \textbf{POPE} & \textbf{MME} & \textbf{MathVista} & \textbf{MathVerse} & \textbf{MMVet} \\ 
    \cline{2-10}
     & val & test & en\_test & test & test & test & testmini & testmini & test \\
    \midrule
    \multicolumn{9}{l}{\textit{Closed-Source Model}} \\
    GPT-4o~\cite{gpt4o} & 69.1 & 54.0 & 83.4 & 58.6 & 85.6 & 2329 & 63.8 & 50.2 & 69.1 \\
    Claude-3.5 Sonnet~\cite{sonnet3_5} & 68.3 & 55.0 & 82.6 & 59.9 & - & 1920 & 67.7 & 41.2 & 70.1 \\
    Gemini-1.5-Pro~\cite{team2023gemini} & 62.2 & 49.4 & 73.9 & 70.4 & 88.2 & - & 63.9 & - & 64.0 \\
    \midrule
    \multicolumn{10}{l}{\textit{Open-Source General Model}} \\ 
    Cambrain-1-8B~\cite{tong2024cambrian} & 42.7 & - & 75.9 & 60.0 & 86.4 & 1803 & 49.0 & - & - \\  
    InternVL2-8B~\cite{internvl2} & 49.3 & 32.5 & 81.7 & 64.4 & 84.2 & 2210 & 58.3 & - & 60.0 \\
    LLaVA-OneVision-7B~\cite{llavaov} & 48.8 & - & - & 66.3 & 88.4 & 1998 & 63.2 & - & 57.5 \\
    MiniCPM-Llama-V-2.5-8B~\cite{minicpm-v} & 45.8 & 19.6 & 77.2 & 63.0 & 86.7 & 2025 & 54.3 & - & - \\
    MiniCPM-V-2.6-8B~\cite{minicpm-v} & 49.8 & 27.2 & 78.0 & 65.0 & 83.2 & 2348 & 60.6 & - & - \\
    IXC-2.5~\cite{ixc25} & 42.9 & - & 82.2 & 67.8 & - & 2229 & 63.8 & - & 51.7 \\
    InternVL2.5-8B~\cite{internvl25} & 56.0 & 38.2 & 84.6 & 70.1 & 90.6 & 2344 & 64.4 & 39.5 & 62.8 \\
    \midrule
    \multicolumn{10}{l}{\textit{Reasoning Model}} \\ 
    LLaVA-CoT-11B~\cite{llava-cot} & - & - & 75.0 & - & - & - & 54.8 & - & 60.3 \\
    LLaVA-Reasoner-8B~\cite{llava-reasoner} & - & - & - & - & - & - & 50.6 & - & - \\
    Insight-V-8B~\cite{insightv} & 50.2 & 24.9 & 82.3 & - & - & 2312 & 59.9 & - & - \\
    Mulberry-7B~\cite{mulberry} & 55.0 & - & - & - & - & 2396 & 63.1 & - & - \\
    Vision-R1-LlamaV-CI-11B~\cite{visionr1} & - & - & - & - & - & 2190 & 62.7 & 27.1 & - \\
    \midrule
    \rowcolor{mygray}
    \multicolumn{10}{l}{\textit{\methodNAME}} \\
    Qwen2.5-VL-7B$^{*}$~\cite{bai2025qwen2.5vl} & 50.3 & 37.7 & 82.6 & 68.6 & 86.7 & 2316 & 68.2 & 46.3 & 61.6 \\
    \methodNAME$\ddag$ & 51.0 & 40.1 & 82.9 & 68.6 & 87.9 & 2307 & 71.2 & 48.8 & 67.5 \\
    \methodNAME & 51.2 & 38.9 & 80.0 & 68.5 & 86.0 & 2400 & 67.5 & 48.0 & 67.1 \\
    \bottomrule
  \end{tabular}
  \end{adjustbox}
  \label{table:generalqa_result}
\end{table*}

\subsection{Data Preparation}
\label{sec:data-prepare}
To enable our model can decide when high resolution is necessary, we collect corresponding VQA samples, including both cases requiring high-resolution images and cases adequately answered using downsampled images.
To achieve this, we use our base policy model, Qwen2.5VL-Instruct, to perform multiple rollouts on the training dataset and classify the samples based on accuracy. Specifically, we set the temperature to 1 and roll out each sample 8 times. If both the high-resolution and downsampled images yield correct answers in all 8 rollouts, we classify the sample as solvable using low resolution. Conversely, if the number of correct answers using the high-resolution image exceeds that of the downsampled image by 6 or more, we classify the sample as requiring high resolution.
By using the above method, we selected 10K samples that require high-resolution images and 10K samples that do not, to train our model.

\section{Experiments}
\subsection{Evaluation Setup}
\mypara{Benchmarks.} 
We evaluate \methodNAME on several general VQA benchmarks, including ChartQA~\cite{masry2022chartqa}, OCRBench~\cite{liu2024ocrbench}, MathVista~\cite{mathvista}, MMVet~\cite{yu2024mmvet}, RealWorldQA~\cite{grok15}, and MathVerse~\cite{zhang2024mathverse}, etc.
Notably, benchmarks such as ChartQA, OCRBench, and MathVista are strongly OCR-related, requiring the model to possess a high level of detail comprehension. The detailed descriptions of these benchmarks are shown in Appendix~\ref{sec:benchmark}.

\mypara{Implementation Details.}
We conduct experiments based on Qwen2.5-VL-7B-Instruct\cite{bai2025qwen2.5vl}. For training, we employ veRL\cite{sheng2024hybridflow} framework and use a total batch size of $512$, with a mini-batch size of $32$, we set the policy LLM learning rate to $1e-6$ and sample $16$ responses per prompt, ensuring a stable and effective training process.
For inference, we use the vLLM framework and set the temperature to 0. Further details are shown in Appendix~\ref{sec:supp-implemantation}.

\subsection{Reinforcement Learning Enables VLM to Be More Effective}
\label{sec:exp-better}

\mypara{Main Results.} 
To demonstrate the effectiveness of our VisionThink, we compare our \methodNAME with the current open-source and closed-source state-of-the-art~(sota) method.
As shown in Table~\ref{table:generalqa_result}, \methodNAME$\ddag$ is used to demonstrate the effectiveness of the LLM-as-Judge strategy on general VQA tasks. It represents a model trained with full image resolution using only accuracy and format rewards, and thus does not incorporate efficiency capabilities.
The results show that our \methodNAME achieves comparable or even superior performance on general VQA tasks while being more efficient. Specifically, MathVerse and MMVet achieve scores of 48.0 and 67.1, representing improvements of 3.7\% and 8.9\%, respectively, over the base model.
Furthermore, our model performs comparably to closed-source models on several benchmarks such as MathVista and MMBench, and even surpasses all closed-source models on MME, achieving a score of 2400.
Besides, as shown in Fig.~\ref{figure:analyze-penalty}(b), by introducing the LLM-as-Judge for test-time scaling, VisionThink's answer outperforms the vanilla model's short direct answer.
Moreover, we scale up the data size to 130K, and further demonstrate the effectiveness of LLM-as-Judge on General VQA Tasks. The results are shown in Appendix~\ref{sec:supp-scaleup}.

\subsection{Reinforcement Learning Enables VLM to Be More Efficient}
\label{sec:exp-faster-smarter}

\begin{figure}[h]
\centering
\includegraphics[width=0.95\textwidth]{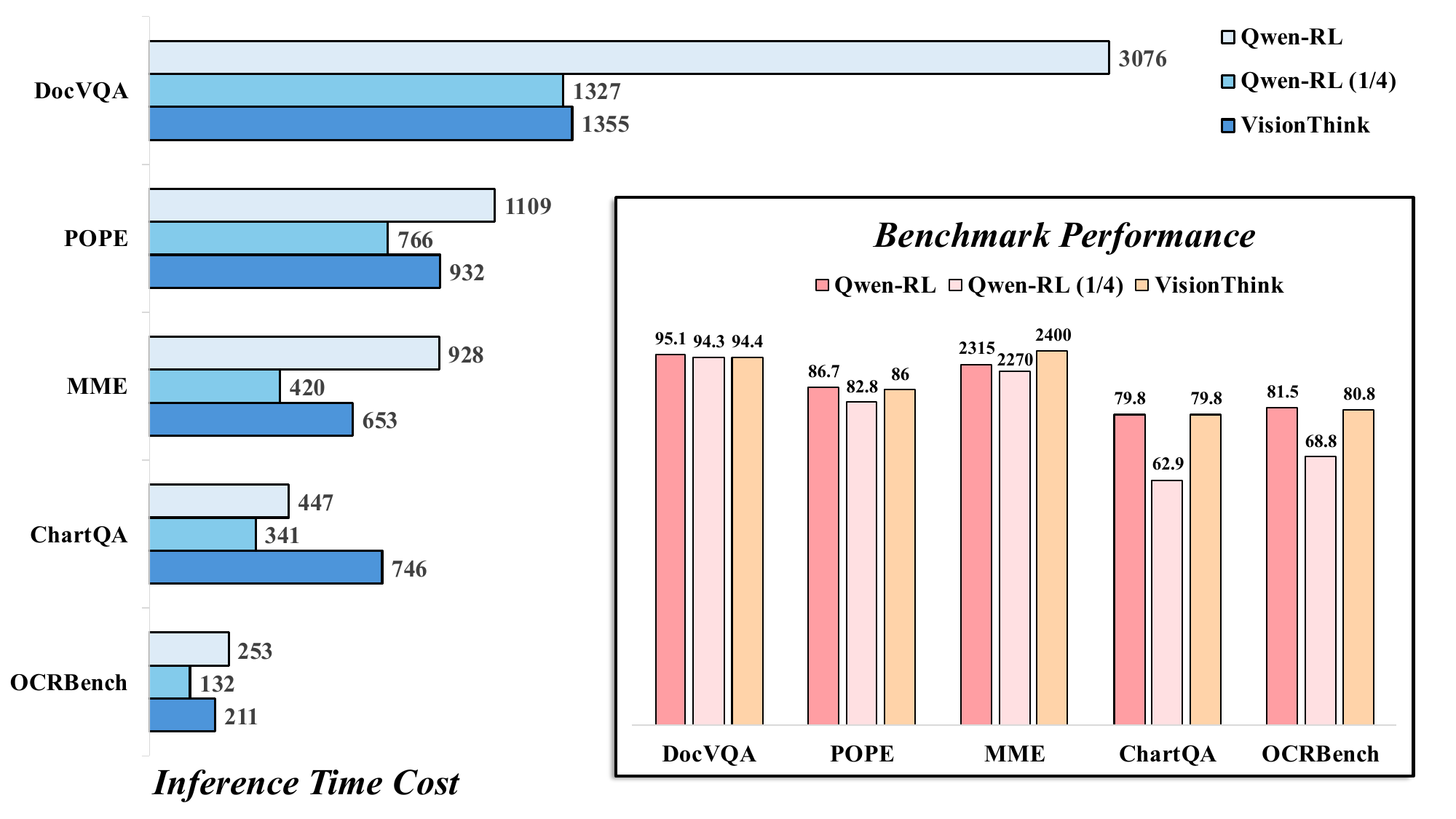}
\caption{\textbf{Inference Time Cost and Benchmark Performance Comparison for Reasoning Model}. Qwen-RL and Qwen-RL~(1/4) represent leveraging the LLM-as-Judge on the Qwen2.5-VL-Instruct Model and inference on full resolution image and 1/4 resolution image, respectively.}
\label{fig:efficiency}
\end{figure}
\mypara{Comparison with the Reasoning Model.}
To demonstrate the efficiency of our model, we first compare our~\methodNAME with QwenRL and QwenRL 1/4, both of which are reasoning models trained using the LLM-as-Judge strategy based on Qwen2.5-VL-7B Instruct. QwenRL and QwenRL 1/4 represent inference using the full-resolution image and the 1/4-resolution image, respectively.
As shown in Fig.~\ref{fig:efficiency}, we compare the inference time costs of the three models. Notably, the reported inference times reflect the actual time consumed during vLLM inference, which we believe best represents efficiency in real-world applications.
The results show that on most benchmarks, our model's inference time is close to that of QwenRL 1/4, which uses 1/4 of the image tokens, and significantly better than the QwenRL model that processes all image tokens.
Specifically, on the DocVQA benchmark, our VisionThink model is more than twice as fast as QwenRL. It also outperforms the baseline by approximately one-third in terms of inference time on benchmarks such as MME and POPE.
It is worth noting that on strongly OCR-dependent benchmarks like ChartQA, our model consumes more time than the baseline QwenRL. This is because~\methodNAME identifies that most questions cannot be answered correctly at low resolution and thus autonomously requests high-resolution images. As a result, the total number of image tokens used by~\methodNAME exceeds that of the baseline, which we consider reasonable. However, such strongly OCR-dependent benchmarks are relatively rare, so the overall efficiency of~\methodNAME remains high.

\mypara{Comparison with the Previous Efficient VLM. }
To further show the effectiveness of our~\methodNAME, we compare it with the previous Efficient VLM method FastV and SparseVLM. Notably, all these methods require computing attention scores to prune visual tokens, which makes them difficult to optimize with FlashAttention2 and may lead to increased memory usage. Furthermore, they are not directly compatible with the efficient inference framework vLLM. Therefore, to ensure a fair comparison, we evaluate model performance while keeping visual token consumption as consistent as possible.
As shown in Table~\ref{table:efficiency_result}, 
our~\methodNAME outperforms previous methods on average across nine benchmarks. Furthermore, previous approaches require a predefined pruning ratio threshold, whereas our method can autonomously decide whether to reduce tokens based on the question and image content. 
As a result, on OCR-Related benchmarks such as ChartQA and OCR Bench, our method significantly surpasses FastV and SparseVLM by 9.0\% and 8.3\%, respectively.

\renewcommand{\multirowsetup}{\centering}
\definecolor{mygray}{gray}{.92}
\definecolor{ForestGreen}{RGB}{34,139,34}
\newcommand{\fg}[1]{\mathbf{\mathcolor{ForestGreen}{#1}}}
\definecolor{Forestred}{RGB}{220,50,50}
\newcommand{\fr}[1]{\mathbf{\mathcolor{Forestred}{#1}}}
\begin{table*}[t]
    \centering
	\caption{\textbf{Comparison with Traditional Efficient VLM Methods.} Vanilla represents the Qwen2.5-VL-7B-Instrcut. The retained ratio of the baseline methods is a predefined hyperparameter, while for \methodNAME, the ratio is determined autonomously by the model and reported as a statistical value. Note that \textit{Down-Sample} refers to the model's performance when directly fed images with their resolution reduced by half. Additional baseline comparison results are shown in Table.~\ref{table:efficiency_supp}}
    
    \begin{adjustbox}{width=\textwidth}
    \setlength{\tabcolsep}{2pt}
    \renewcommand{\arraystretch}{1.4}
    \footnotesize
	\centering
    \begin{tabular}{p{3.5cm}|c c c c c c c c c| c}
        \toprule
        \multirow{2}{*}{\textbf{Method}} & \textbf{ChartQA$^{\dagger}$} & \textbf{OCRBench} & \textbf{DocVQA} & \textbf{MME} & \textbf{MMVet} & \textbf{RealWorldQA} & \textbf{POPE} & \textbf{MathVista} & \textbf{MathVerse} & \multirow{2}{*}{\makecell[c]{\textbf{Avg}.}}\\
        \cline{2-10}
        & test & test & val & test & test & test & test & testmini & testmini &  \\
        \midrule
        \rowcolor{mygray}
        \multicolumn{11}{c}{\textit{Retain 100\% Visual Tokens Across All Benchmarks}}\\
        \multirow{2}{*}{Vanilla} & 79.8 & 81.5 & 95.1 & 2316 & 61.6 & 68.6 & 86.7 & 68.2 & 46.3 & \multirow{2}*{100\%} \\
        ~ & 100\% & 100\% & 100\% & 100\% & 100\% & 100\% & 100\% & 100\% & 100\%  & ~ \\
        \hline
        \rowcolor{mygray}
        \multicolumn{11}{c}{\textit{Retain 25\% Visual Tokens Across All Benchmarks}}\\
        \multirow{2}{*}{Down-Sample} & 62.9 & 68.8 & 94.3 & 2270 & 54.5 & 68.8 & 82.8 & 62.2 & 43.1 & \multirow{2}*{92.1\%} \\
        ~ & 78.8\% & 84.4\% & 99.1\% & 98.0\% & 88.5\% & 100.3\% & 95.5\% & 91.2\% & 93.1\%  & ~ \\
        \hline
        \rowcolor{mygray}
        \multicolumn{11}{c}{\textit{Retain 50\% Visual Tokens Across All Benchmarks}} \\
        \multirow{2}{*}{SparseVLM~(ICML 2025)} & 73.2 & 75.6 & 66.8 & 2282 & 51.5  & 68.4 & 85.5 & 66.6 & 45.1 & \multirow{2}*{\textbf{92.2\%}} \\
        ~ & 91.7\% & 92.7\% & 70.2\% & 98.5\% & 83.6\% & 99.7\% & 98.6\% & 97.6\% & 97.4\% &  ~ \\
        \hline
        \multirow{2}{*}{FastV~(ECCV 2024)} & 72.6 & 75.8 & 93.6 & 2308 & 52.8 & 68.8 & 84.7 & 63.7 & 45.0 & \multirow{2}*{\textbf{95.8\%}} \\
        ~ & 91.0\% & 93.0\% & 98.4\% & 99.6\% & 85.7\% & 100.3\% & 97.7\% & 93.4\% & 97.2\% & ~\\
        \hline
        \rowcolor{mygray}
        \multicolumn{11}{c}{\textit{Retain 70\% Visual Tokens Across All Benchmarks}} \\
        \multirow{2}{*}{SparseVLM~(ICML 2025)} & 75.8 & 79.3 & 68.7 & 2276 & 53.7 & 68.5 & 85.4 & 66.3 & 45.1 & \multirow{2}*{\textbf{93.6\%}} \\
        ~ & 94.9\% & 97.3\% & 72.2\% & 98.3\% & 87.2\% & 99.8\% & 98.5\% & 97.2\% & 97.4\% &  ~ \\
        \hline
        \multirow{2}{*}{FastV~(ECCV 2024)} & 77.2 & 82.2 & 94.4 & 2342 & 56.0 & 68.6 & 85.9 & 65.9 & 46.9 & \multirow{2}*{\textbf{98.4\%}} \\
        ~ & 96.7\% & 100.8\% & 99.3\% & 101.1\% & 90.9\% & 100\% & 99.1\% & 96.6\% & 101.3\% & ~\\
        \hline
        \rowcolor{mygray}
        \multicolumn{11}{c}{\textit{Retain Approximately 51.3\% Visual Tokens Across All Benchmarks}} \\
        \multirow{2}{*}{\methodNAME} & 79.8 & 80.8 & 94.4 & 2400 & 68.5 & 67.1 & 86.0 & 67.5 & 48.0 & \multirow{2}*{$\fr{101.4\%}$} \\
        ~ & 100\% & 99.1\% & 99.3\% & 103.6\% & 111.2\% & 97.8\% & 99.2\% & 99.0\% & 103.7\% &  ~ \\
        \bottomrule
	\end{tabular}
      \end{adjustbox}

	\label{table:efficiency_result}
    \vspace{-0.5cm}
\end{table*}

\subsection{Reinforcement Learning Enables VLM to Be Smarter}

\begin{figure}[h]
\centering
\includegraphics[width=0.98\textwidth]{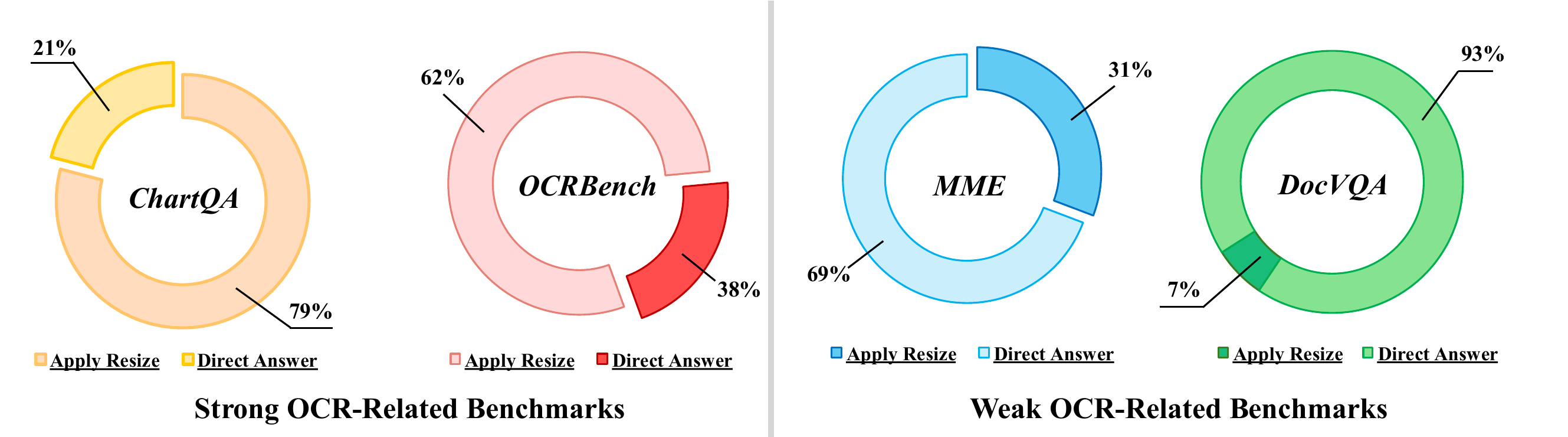}
\caption{\textbf{\methodNAME smartly determine the high-resolution image ratio.} Apply Resize indicates that the model autonomously requests to view the original high-resolution image, while Direct Answer indicates that the model is able to answer the question using only the 1/4-sized image.}
\label{figure:smart-ratio}
\end{figure}

In this section, we present the proportion of samples across different benchmarks for which our \methodNAME gives direct answers versus those for which it requests high-resolution images. This illustrates the model’s ability to smartly determine whether the information in the downsampled image is sufficient.
As shown in Fig.~\ref{figure:smart-ratio}, we observe that on benchmarks such as ChartQA and OCRBench, which require detailed visual understanding, our model shows a higher ratio of requests for high-resolution images. In contrast, for benchmarks like MME and DocVQA, at least 70\% of the samples can be answered directly using low-resolution images at 1/4 of the original resolution.
These results align with human intuition: most daily questions do not require high-resolution images, and only OCR-related tasks truly depend on them.
Furthermore, to better demonstrate the `smart' capabilities of \methodNAME, we conduct case studies in Appendix~\ref{sec:supp-qualitative}.

\subsection{Relationship of the EfficientVLM methods and \methodNAME.}
\mypara{Key Differences. }
Traditional EfficientVLM methods take a redundant image as input and attempt to remove the redundancy during inference. However, this process typically relies on fixed thresholds, which may yield acceptable performance on standard VQA tasks but result in poor performance on OCR-related or detail-sensitive scenarios, limiting their practical applicability.
 In contrast, \methodNAME inputs compressed visual tokens and enables the model to autonomously determine whether a higher-resolution image is needed. Ideally, this approach avoids any performance degradation.
 
\mypara{Integration Potential. }
Our proposed \methodNAME essentially introduces a new paradigm for reading images, which can be integrated with existing Efficient VLMs. 
In this paper, to provide a straightforward validation of \methodNAME, we chose to use image resizing perform token compression.  We believe that adopting more advanced token compression techniques could further improve the model’s direct answering accuracy, consequently, enhance its overall efficiency.
Further discussions are shown in Appendix~\ref{sec:supp-discuss}.

\section{Related Works}
\mypara{Vision Language Model Reasoning. }
With the advancement of LLM reasoning capabilities~\cite{guo2025deepseek}, many studies have aimed to improve the reasoning abilities of VLMs~\cite{zhang2024improve,mitra2024compositional,luan2024textcot}. One common approach is using Chain-of-Thought (CoT) prompting to construct SFT datasets. However, the CoTs generated often lack natural human cognitive processes, limiting their effectiveness and generalization.
Furthermore, inspired by DeepSeek-R1~\cite{guo2025deepseek}, some studies have attempted to transfer this reasoning paradigm to vision tasks~\cite{yao2024mulberry,thawakar2025llamav,huang2025vision,yang2025r1}. However, most current approaches remain limited to the visual math and fail to generalize to general VQA tasks.
In contrast,~\methodNAME successfully applies effective reinforcement learning to general VQA by leveraging the LLM-as-Judge strategy.
Due to space limitations, additional related work on efficient VLMs and LLM-based reasoning is presented in Appendix~\ref{sec:appendix-related}.

\section{Concluding Remarks}
\subsection{Summary}
In this work, we introduce \methodNAME, a novel paradigm for General VQA that enhances efficiency and performance. By initially processing a downsampled image and using reinforcement learning to selectively upscale to higher resolution when needed, \methodNAME optimizes computational resources while preserving accuracy. Leveraging the LLM-as-Judge strategy and a tailored reward function, our approach outperforms prior state-of-the-art models across diverse VQA benchmarks, particularly in tasks requiring fine-grained details like OCR. We believe \methodNAME demonstrates the potential of reinforcement learning in vision-language models and encourages the development of more effective and efficient AI systems. 

\subsection{Limitations and Future work}
In this work, we focus on the setting of 2x resolution upscaling and at most two turns of conversations and yield promising results. However, it has not been extended to the setting of flexible resolution upscaling. Besides, incorporating more visual tools such as cropping would further bring benefits in both efficiency and performance. Furthermore, multi-turn (for example, more than 5 turns) image tool calls could gain more in solving complex visual problems. 

Additionally, our paper utilizes image resizing to reduce the number of visual tokens. This simple method achieves a good balance between performance and efficiency via reinforcement learning. We hope this work inspires further research in the field of efficient reasoning vision language models, especially on making models smarter and more human-like.
We will continue to explore the path toward building more general, powerful, and efficient vision-language models.
\clearpage
\bibliography{main}

\begin{thebibliography}{10}

\bibitem{achiam2023gpt}
Josh Achiam, Steven Adler, Sandhini Agarwal, Lama Ahmad, Ilge Akkaya, Florencia~Leoni Aleman, Diogo Almeida, Janko Altenschmidt, Sam Altman, Shyamal Anadkat, et~al.
\newblock Gpt-4 technical report.
\newblock {\em arXiv:2303.08774}, 2023.

\bibitem{sonnet3_5}
Anthropic.
\newblock Claude 3.5 sonnet, 2024.

\bibitem{Qwen-VL}
Jinze Bai, Shuai Bai, Shusheng Yang, Shijie Wang, Sinan Tan, Peng Wang, Junyang Lin, Chang Zhou, and Jingren Zhou.
\newblock Qwen-{VL}: A frontier large vision-language model with versatile abilities.
\newblock {\em arXiv:2308.12966}, 2023.

\bibitem{bai2023qwen}
Jinze Bai, Shuai Bai, Shusheng Yang, Shijie Wang, Sinan Tan, Peng Wang, Junyang Lin, Chang Zhou, and Jingren Zhou.
\newblock Qwen-vl: A frontier large vision-language model with versatile abilities.
\newblock {\em arXiv preprint arXiv:2308.12966}, 2023.

\bibitem{bai2025qwen2.5vl}
Shuai Bai, Keqin Chen, Xuejing Liu, Jialin Wang, Wenbin Ge, Sibo Song, Kai Dang, Peng Wang, Shijie Wang, Jun Tang, et~al.
\newblock Qwen2. 5-vl technical report.
\newblock {\em arXiv preprint arXiv:2502.13923}, 2025.

\bibitem{Qwen2.5-VL}
Shuai Bai, Keqin Chen, Xuejing Liu, Jialin Wang, Wenbin Ge, Sibo Song, Kai Dang, Peng Wang, Shijie Wang, Jun Tang, Humen Zhong, Yuanzhi Zhu, Mingkun Yang, Zhaohai Li, Jianqiang Wan, Pengfei Wang, Wei Ding, Zheren Fu, Yiheng Xu, Jiabo Ye, Xi~Zhang, Tianbao Xie, Zesen Cheng, Hang Zhang, Zhibo Yang, Haiyang Xu, and Junyang Lin.
\newblock Qwen2.5-vl technical report.
\newblock {\em arXiv preprint arXiv:2502.13923}, 2025.

\bibitem{besta2024graph}
Maciej Besta, Nils Blach, Ales Kubicek, Robert Gerstenberger, Michal Podstawski, Lukas Gianinazzi, Joanna Gajda, Tomasz Lehmann, Hubert Niewiadomski, Piotr Nyczyk, et~al.
\newblock Graph of thoughts: Solving elaborate problems with large language models.
\newblock In {\em Proceedings of the AAAI Conference on Artificial Intelligence}, volume~38, pages 17682--17690, 2024.

\bibitem{chen2024image}
Liang Chen, Haozhe Zhao, Tianyu Liu, Shuai Bai, Junyang Lin, Chang Zhou, and Baobao Chang.
\newblock An image is worth 1/2 tokens after layer 2: Plug-and-play inference acceleration for large vision-language models, 2024.

\bibitem{chen2023sharegpt4v}
Lin Chen, Jisong Li, Xiaoyi Dong, Pan Zhang, Conghui He, Jiaqi Wang, Feng Zhao, and Dahua Lin.
\newblock Sharegpt4v: Improving large multi-modal models with better captions.
\newblock {\em arXiv:2311.12793}, 2023.

\bibitem{chen2023programthoughtspromptingdisentangling}
Wenhu Chen, Xueguang Ma, Xinyi Wang, and William~W. Cohen.
\newblock Program of thoughts prompting: Disentangling computation from reasoning for numerical reasoning tasks, 2023.

\bibitem{internvl25}
Zhe Chen, Weiyun Wang, Yue Cao, Yangzhou Liu, Zhangwei Gao, Erfei Cui, Jinguo Zhu, Shenglong Ye, Hao Tian, Zhaoyang Liu, et~al.
\newblock Expanding performance boundaries of open-source multimodal models with model, data, and test-time scaling.
\newblock {\em arXiv preprint arXiv:2412.05271}, 2024.

\bibitem{internvl2}
Zhe Chen, Weiyun Wang, Hao Tian, Shenglong Ye, Zhangwei Gao, Erfei Cui, Wenwen Tong, Kongzhi Hu, Jiapeng Luo, Zheng Ma, et~al.
\newblock Internvl2: Better than the best—expanding performance boundaries of open-source multimodal models with the progressive scaling strategy, 2024.

\bibitem{chiang2023vicuna}
Wei-Lin Chiang, Zhuohan Li, Zi~Lin, Ying Sheng, Zhanghao Wu, Hao Zhang, Lianmin Zheng, Siyuan Zhuang, Yonghao Zhuang, Joseph~E Gonzalez, et~al.
\newblock Vicuna: An open-source chatbot impressing gpt-4 with 90\%* chatgpt quality.
\newblock {\em See https://vicuna. lmsys. org (accessed 14 April 2023)}, 2(3):6, 2023.

\bibitem{insightv}
Yuhao Dong, Zuyan Liu, Hai-Long Sun, Jingkang Yang, Winston Hu, Yongming Rao, and Ziwei Liu.
\newblock Insight-v: Exploring long-chain visual reasoning with multimodal large language models.
\newblock {\em arXiv preprint arXiv:2411.14432}, 2024.

\bibitem{fu2023mme}
Chaoyou Fu, Peixian Chen, Yunhang Shen, Yulei Qin, Mengdan Zhang, Xu~Lin, Zhenyu Qiu, Wei Lin, Jinrui Yang, Xiawu Zheng, et~al.
\newblock Mme: A comprehensive evaluation benchmark for multimodal large language models.
\newblock {\em arXiv:2306.13394}, 2023.

\bibitem{guo2025deepseekr1}
Daya Guo, Dejian Yang, Haowei Zhang, Junxiao Song, Ruoyu Zhang, Runxin Xu, Qihao Zhu, Shirong Ma, Peiyi Wang, Xiao Bi, et~al.
\newblock Deepseek-r1: Incentivizing reasoning capability in llms via reinforcement learning.
\newblock {\em arXiv preprint arXiv:2501.12948}, 2025.

\bibitem{guo2025deepseek}
Daya Guo, Dejian Yang, Haowei Zhang, Junxiao Song, Ruoyu Zhang, Runxin Xu, Qihao Zhu, Shirong Ma, Peiyi Wang, Xiao Bi, et~al.
\newblock Deepseek-r1: Incentivizing reasoning capability in llms via reinforcement learning.
\newblock {\em arXiv preprint arXiv:2501.12948}, 2025.

\bibitem{he2024zipvl}
Yefei He, Feng Chen, Jing Liu, Wenqi Shao, Hong Zhou, Kaipeng Zhang, and Bohan Zhuang.
\newblock Zipvl: Efficient large vision-language models with dynamic token sparsification and kv cache compression.
\newblock {\em arXiv preprint arXiv:2410.08584}, 2024.

\bibitem{visionr1}
Wenxuan Huang, Bohan Jia, Zijie Zhai, Shaosheng Cao, Zheyu Ye, Fei Zhao, Zhe Xu, Yao Hu, and Shaohui Lin.
\newblock Vision-r1: Incentivizing reasoning capability in multimodal large language models.
\newblock {\em arXiv preprint arXiv:2503.06749}, 2025.

\bibitem{huang2025vision}
Wenxuan Huang, Bohan Jia, Zijie Zhai, Shaosheng Cao, Zheyu Ye, Fei Zhao, Zhe Xu, Yao Hu, and Shaohui Lin.
\newblock Vision-r1: Incentivizing reasoning capability in multimodal large language models.
\newblock {\em arXiv preprint arXiv:2503.06749}, 2025.

\bibitem{jaech2024openai}
Aaron Jaech, Adam Kalai, Adam Lerer, Adam Richardson, Ahmed El-Kishky, Aiden Low, Alec Helyar, Aleksander Madry, Alex Beutel, Alex Carney, et~al.
\newblock Openai o1 system card.
\newblock {\em arXiv preprint arXiv:2412.16720}, 2024.

\bibitem{jian2023expedited}
Yiren Jian, Tingkai Liu, Yunzhe Tao, Chunhui Zhang, Soroush Vosoughi, and Hongxia Yang.
\newblock Expedited training of visual conditioned language generation via redundancy reduction.
\newblock {\em arXiv preprint arXiv:2310.03291}, 2023.

\bibitem{jin2025search}
Bowen Jin, Hansi Zeng, Zhenrui Yue, Jinsung Yoon, Sercan Arik, Dong Wang, Hamed Zamani, and Jiawei Han.
\newblock Search-r1: Training llms to reason and leverage search engines with reinforcement learning.
\newblock {\em arXiv preprint arXiv:2503.09516}, 2025.

\bibitem{kim24openvla}
{Moo Jin} Kim, Karl Pertsch, Siddharth Karamcheti, Ted Xiao, Ashwin Balakrishna, Suraj Nair, Rafael Rafailov, Ethan Foster, Grace Lam, Pannag Sanketi, Quan Vuong, Thomas Kollar, Benjamin Burchfiel, Russ Tedrake, Dorsa Sadigh, Sergey Levine, Percy Liang, and Chelsea Finn.
\newblock Openvla: An open-source vision-language-action model.
\newblock {\em arXiv preprint arXiv:2406.09246}, 2024.

\bibitem{vllm}
Woosuk Kwon, Zhuohan Li, Siyuan Zhuang, Ying Sheng, Lianmin Zheng, Cody~Hao Yu, Joseph~E. Gonzalez, Hao Zhang, and Ion Stoica.
\newblock Efficient memory management for large language model serving with pagedattention.
\newblock In {\em Proceedings of the ACM SIGOPS 29th Symposium on Operating Systems Principles}, 2023.

\bibitem{lai2023lisa}
Xin Lai, Zhuotao Tian, Yukang Chen, Yanwei Li, Yuhui Yuan, Shu Liu, and Jiaya Jia.
\newblock Lisa: Reasoning segmentation via large language model.
\newblock {\em arXiv preprint arXiv:2308.00692}, 2023.

\bibitem{lai2024stepdpostepwisepreferenceoptimization}
Xin Lai, Zhuotao Tian, Yukang Chen, Senqiao Yang, Xiangru Peng, and Jiaya Jia.
\newblock Step-dpo: Step-wise preference optimization for long-chain reasoning of llms, 2024.

\bibitem{llavaov}
Bo~Li, Yuanhan Zhang, Dong Guo, Renrui Zhang, Feng Li, Hao Zhang, Kaichen Zhang, Peiyuan Zhang, Yanwei Li, Ziwei Liu, et~al.
\newblock Llava-onevision: Easy visual task transfer.
\newblock {\em arXiv preprint arXiv:2408.03326}, 2024.

\bibitem{li2025logits}
Jingyao Li, Senqiao Yang, Sitong Wu, Han Shi, Chuanyang Zheng, Hong Xu, and Jiaya Jia.
\newblock Logits-based finetuning.
\newblock {\em arXiv preprint arXiv:2505.24461}, 2025.

\bibitem{li2023blip}
Junnan Li, Dongxu Li, Silvio Savarese, and Steven Hoi.
\newblock Blip-2: Bootstrapping language-image pre-training with frozen image encoders and large language models.
\newblock In {\em International conference on machine learning}, 2023.

\bibitem{li2024mini}
Yanwei Li, Yuechen Zhang, Chengyao Wang, Zhisheng Zhong, Yixin Chen, Ruihang Chu, Shaoteng Liu, and Jiaya Jia.
\newblock Mini-gemini: Mining the potential of multi-modality vision language models.
\newblock {\em arXiv:2403.18814}, 2024.

\bibitem{li2023evaluating}
Yifan Li, Yifan Du, Kun Zhou, Jinpeng Wang, Wayne~Xin Zhao, and Ji-Rong Wen.
\newblock Evaluating object hallucination in large vision-language models.
\newblock {\em arXiv:2305.10355}, 2023.

\bibitem{liu2023improvedllava}
Haotian Liu, Chunyuan Li, Yuheng Li, and Yong~Jae Lee.
\newblock Improved baselines with visual instruction tuning.
\newblock {\em arXiv:2310.03744}, 2023.

\bibitem{liu2024llavanext}
Haotian Liu, Chunyuan Li, Yuheng Li, Bo~Li, Yuanhan Zhang, Sheng Shen, and Yong~Jae Lee.
\newblock Llava-next: Improved reasoning, ocr, and world knowledge, January 2024.

\bibitem{liu2024visual}
Haotian Liu, Chunyuan Li, Qingyang Wu, and Yong~Jae Lee.
\newblock Visual instruction tuning.
\newblock {\em Advances in neural information processing systems}, 2024.

\bibitem{liu2024robomamba}
Jiaming Liu, Mengzhen Liu, Zhenyu Wang, Lily Lee, Kaichen Zhou, Pengju An, Senqiao Yang, Renrui Zhang, Yandong Guo, and Shanghang Zhang.
\newblock Robomamba: Multimodal state space model for efficient robot reasoning and manipulation.
\newblock {\em arXiv preprint arXiv:2406.04339}, 2024.

\bibitem{liu2024ocrbench}
Yuliang Liu, Zhang Li, Mingxin Huang, Biao Yang, Wenwen Yu, Chunyuan Li, Xucheng Yin, Cheng lin Liu, Lianwen Jin, and Xiang Bai.
\newblock Ocrbench: On the hidden mystery of ocr in large multimodal models.
\newblock {\em arXiv:2305.07895}, 2023.

\bibitem{liu2025seg}
Yuqi Liu, Bohao Peng, Zhisheng Zhong, Zihao Yue, Fanbin Lu, Bei Yu, and Jiaya Jia.
\newblock Seg-zero: Reasoning-chain guided segmentation via cognitive reinforcement.
\newblock {\em arXiv preprint arXiv:2503.06520}, 2025.

\bibitem{liu2025visual}
Ziyu Liu, Zeyi Sun, Yuhang Zang, Xiaoyi Dong, Yuhang Cao, Haodong Duan, Dahua Lin, and Jiaqi Wang.
\newblock Visual-rft: Visual reinforcement fine-tuning.
\newblock {\em arXiv preprint arXiv:2503.01785}, 2025.

\bibitem{mathvista}
Pan Lu, Hritik Bansal, Tony Xia, Jiacheng Liu, Chunyuan Li, Hannaneh Hajishirzi, Hao Cheng, Kai{-}Wei Chang, Michel Galley, and Jianfeng Gao.
\newblock Mathvista: Evaluating mathematical reasoning of foundation models in visual contexts.
\newblock In {\em ICLR}, 2024.

\bibitem{luan2024textcot}
Bozhi Luan, Hao Feng, Hong Chen, Yonghui Wang, Wengang Zhou, and Houqiang Li.
\newblock Textcot: Zoom in for enhanced multimodal text-rich image understanding.
\newblock {\em arXiv preprint arXiv:2404.09797}, 2024.

\bibitem{luong2024reftreasoningreinforcedfinetuning}
Trung~Quoc Luong, Xinbo Zhang, Zhanming Jie, Peng Sun, Xiaoran Jin, and Hang Li.
\newblock Reft: Reasoning with reinforced fine-tuning, 2024.

\bibitem{masry2022chartqa}
Ahmed Masry, Do~Xuan Long, Jia~Qing Tan, Shafiq Joty, and Enamul Hoque.
\newblock Chartqa: A benchmark for question answering about charts with visual and logical reasoning.
\newblock {\em arXiv preprint arXiv:2203.10244}, 2022.

\bibitem{docvqa}
Minesh Mathew, Dimosthenis Karatzas, and CV~Jawahar.
\newblock Docvqa: A dataset for vqa on document images.
\newblock In {\em WACV}, 2021.

\bibitem{meng2025mm}
Fanqing Meng, Lingxiao Du, Zongkai Liu, Zhixiang Zhou, Quanfeng Lu, Daocheng Fu, Tiancheng Han, Botian Shi, Wenhai Wang, Junjun He, et~al.
\newblock Mm-eureka: Exploring the frontiers of multimodal reasoning with rule-based reinforcement learning.
\newblock {\em arXiv preprint arXiv:2503.07365}, 2025.

\bibitem{mitra2024compositional}
Chancharik Mitra, Brandon Huang, Trevor Darrell, and Roei Herzig.
\newblock Compositional chain-of-thought prompting for large multimodal models.
\newblock In {\em Proceedings of the IEEE/CVF Conference on Computer Vision and Pattern Recognition}, pages 14420--14431, 2024.

\bibitem{openai2024openaio1card}
OpenAI, :, Aaron Jaech, Adam Kalai, Adam Lerer, Adam Richardson, Ahmed El-Kishky, Aiden Low, Alec Helyar, Aleksander Madry, Alex Beutel, Alex Carney, Alex Iftimie, Alex Karpenko, Alex~Tachard Passos, Alexander Neitz, Alexander Prokofiev, Alexander Wei, Allison Tam, Ally Bennett, Ananya Kumar, Andre Saraiva, Andrea Vallone, Andrew Duberstein, Andrew Kondrich, Andrey Mishchenko, Andy Applebaum, Angela Jiang, Ashvin Nair, Barret Zoph, Behrooz Ghorbani, Ben Rossen, Benjamin Sokolowsky, Boaz Barak, Bob McGrew, Borys Minaiev, Botao Hao, Bowen Baker, Brandon Houghton, Brandon McKinzie, Brydon Eastman, Camillo Lugaresi, Cary Bassin, Cary Hudson, Chak~Ming Li, Charles de~Bourcy, Chelsea Voss, Chen Shen, Chong Zhang, Chris Koch, Chris Orsinger, Christopher Hesse, Claudia Fischer, Clive Chan, Dan Roberts, Daniel Kappler, Daniel Levy, Daniel Selsam, David Dohan, David Farhi, David Mely, David Robinson, Dimitris Tsipras, Doug Li, Dragos Oprica, Eben Freeman, Eddie Zhang, Edmund Wong, Elizabeth Proehl, Enoch Cheung, Eric
  Mitchell, Eric Wallace, Erik Ritter, Evan Mays, Fan Wang, Felipe~Petroski Such, Filippo Raso, Florencia Leoni, Foivos Tsimpourlas, Francis Song, Fred von Lohmann, Freddie Sulit, Geoff Salmon, Giambattista Parascandolo, Gildas Chabot, Grace Zhao, Greg Brockman, Guillaume Leclerc, Hadi Salman, Haiming Bao, Hao Sheng, Hart Andrin, Hessam Bagherinezhad, Hongyu Ren, Hunter Lightman, Hyung~Won Chung, Ian Kivlichan, Ian O'Connell, Ian Osband, Ignasi~Clavera Gilaberte, Ilge Akkaya, Ilya Kostrikov, Ilya Sutskever, Irina Kofman, Jakub Pachocki, James Lennon, Jason Wei, Jean Harb, Jerry Twore, Jiacheng Feng, Jiahui Yu, Jiayi Weng, Jie Tang, Jieqi Yu, Joaquin~Quiñonero Candela, Joe Palermo, Joel Parish, Johannes Heidecke, John Hallman, John Rizzo, Jonathan Gordon, Jonathan Uesato, Jonathan Ward, Joost Huizinga, Julie Wang, Kai Chen, Kai Xiao, Karan Singhal, Karina Nguyen, Karl Cobbe, Katy Shi, Kayla Wood, Kendra Rimbach, Keren Gu-Lemberg, Kevin Liu, Kevin Lu, Kevin Stone, Kevin Yu, Lama Ahmad, Lauren Yang, Leo Liu,
  Leon Maksin, Leyton Ho, Liam Fedus, Lilian Weng, Linden Li, Lindsay McCallum, Lindsey Held, Lorenz Kuhn, Lukas Kondraciuk, Lukasz Kaiser, Luke Metz, Madelaine Boyd, Maja Trebacz, Manas Joglekar, Mark Chen, Marko Tintor, Mason Meyer, Matt Jones, Matt Kaufer, Max Schwarzer, Meghan Shah, Mehmet Yatbaz, Melody~Y. Guan, Mengyuan Xu, Mengyuan Yan, Mia Glaese, Mianna Chen, Michael Lampe, Michael Malek, Michele Wang, Michelle Fradin, Mike McClay, Mikhail Pavlov, Miles Wang, Mingxuan Wang, Mira Murati, Mo~Bavarian, Mostafa Rohaninejad, Nat McAleese, Neil Chowdhury, Neil Chowdhury, Nick Ryder, Nikolas Tezak, Noam Brown, Ofir Nachum, Oleg Boiko, Oleg Murk, Olivia Watkins, Patrick Chao, Paul Ashbourne, Pavel Izmailov, Peter Zhokhov, Rachel Dias, Rahul Arora, Randall Lin, Rapha~Gontijo Lopes, Raz Gaon, Reah Miyara, Reimar Leike, Renny Hwang, Rhythm Garg, Robin Brown, Roshan James, Rui Shu, Ryan Cheu, Ryan Greene, Saachi Jain, Sam Altman, Sam Toizer, Sam Toyer, Samuel Miserendino, Sandhini Agarwal, Santiago Hernandez,
  Sasha Baker, Scott McKinney, Scottie Yan, Shengjia Zhao, Shengli Hu, Shibani Santurkar, Shraman~Ray Chaudhuri, Shuyuan Zhang, Siyuan Fu, Spencer Papay, Steph Lin, Suchir Balaji, Suvansh Sanjeev, Szymon Sidor, Tal Broda, Aidan Clark, Tao Wang, Taylor Gordon, Ted Sanders, Tejal Patwardhan, Thibault Sottiaux, Thomas Degry, Thomas Dimson, Tianhao Zheng, Timur Garipov, Tom Stasi, Trapit Bansal, Trevor Creech, Troy Peterson, Tyna Eloundou, Valerie Qi, Vineet Kosaraju, Vinnie Monaco, Vitchyr Pong, Vlad Fomenko, Weiyi Zheng, Wenda Zhou, Wes McCabe, Wojciech Zaremba, Yann Dubois, Yinghai Lu, Yining Chen, Young Cha, Yu~Bai, Yuchen He, Yuchen Zhang, Yunyun Wang, Zheng Shao, and Zhuohan Li.
\newblock Openai o1 system card, 2024.

\bibitem{gpt4o}
OpenAI.
\newblock Hello gpt-4o, 2024.

\bibitem{peng2023instruction}
Baolin Peng, Chunyuan Li, Pengcheng He, Michel Galley, and Jianfeng Gao.
\newblock Instruction tuning with gpt-4.
\newblock {\em arXiv:2304.03277}, 2023.

\bibitem{qu2024mobile}
Guanqiao Qu, Qiyuan Chen, Wei Wei, Zheng Lin, Xianhao Chen, and Kaibin Huang.
\newblock Mobile edge intelligence for large language models: A contemporary survey.
\newblock {\em arXiv preprint arXiv:2407.18921}, 2024.

\bibitem{qu2025does}
Tianyuan Qu, Longxiang Tang, Bohao Peng, Senqiao Yang, Bei Yu, and Jiaya Jia.
\newblock Does your vision-language model get lost in the long video sampling dilemma?
\newblock {\em arXiv preprint arXiv:2503.12496}, 2025.

\bibitem{shao2024deepseekmath}
Zhihong Shao, Peiyi Wang, Qihao Zhu, Runxin Xu, Junxiao Song, Xiao Bi, Haowei Zhang, Mingchuan Zhang, YK~Li, Y~Wu, et~al.
\newblock Deepseekmath: Pushing the limits of mathematical reasoning in open language models.
\newblock {\em arXiv preprint arXiv:2402.03300}, 2024.

\bibitem{shen2025vlm}
Haozhan Shen, Peng Liu, Jingcheng Li, Chunxin Fang, Yibo Ma, Jiajia Liao, Qiaoli Shen, Zilun Zhang, Kangjia Zhao, Qianqian Zhang, et~al.
\newblock Vlm-r1: A stable and generalizable r1-style large vision-language model.
\newblock {\em arXiv preprint arXiv:2504.07615}, 2025.

\bibitem{sheng2024hybridflow}
Guangming Sheng, Chi Zhang, Zilingfeng Ye, Xibin Wu, Wang Zhang, Ru~Zhang, Yanghua Peng, Haibin Lin, and Chuan Wu.
\newblock Hybridflow: A flexible and efficient rlhf framework.
\newblock {\em arXiv preprint arXiv: 2409.19256}, 2024.

\bibitem{shi2023upop}
Dachuan Shi, Chaofan Tao, Ying Jin, Zhendong Yang, Chun Yuan, and Jiaqi Wang.
\newblock Upop: Unified and progressive pruning for compressing vision-language transformers.
\newblock In {\em International Conference on Machine Learning}, pages 31292--31311. PMLR, 2023.

\bibitem{snell2024scalingllmtesttimecompute}
Charlie Snell, Jaehoon Lee, Kelvin Xu, and Aviral Kumar.
\newblock Scaling llm test-time compute optimally can be more effective than scaling model parameters, 2024.

\bibitem{team2023gemini}
Gemini Team, Rohan Anil, Sebastian Borgeaud, Jean-Baptiste Alayrac, Jiahui Yu, Radu Soricut, Johan Schalkwyk, Andrew~M Dai, Anja Hauth, Katie Millican, et~al.
\newblock Gemini: a family of highly capable multimodal models.
\newblock {\em arXiv preprint arXiv:2312.11805}, 2023.

\bibitem{kimiteam2025kimik15scalingreinforcement}
Kimi Team, Angang Du, Bofei Gao, Bowei Xing, Changjiu Jiang, Cheng Chen, Cheng Li, Chenjun Xiao, Chenzhuang Du, Chonghua Liao, Chuning Tang, Congcong Wang, Dehao Zhang, Enming Yuan, Enzhe Lu, Fengxiang Tang, Flood Sung, Guangda Wei, Guokun Lai, Haiqing Guo, Han Zhu, Hao Ding, Hao Hu, Hao Yang, Hao Zhang, Haotian Yao, Haotian Zhao, Haoyu Lu, Haoze Li, Haozhen Yu, Hongcheng Gao, Huabin Zheng, Huan Yuan, Jia Chen, Jianhang Guo, Jianlin Su, Jianzhou Wang, Jie Zhao, Jin Zhang, Jingyuan Liu, Junjie Yan, Junyan Wu, Lidong Shi, Ling Ye, Longhui Yu, Mengnan Dong, Neo Zhang, Ningchen Ma, Qiwei Pan, Qucheng Gong, Shaowei Liu, Shengling Ma, Shupeng Wei, Sihan Cao, Siying Huang, Tao Jiang, Weihao Gao, Weimin Xiong, Weiran He, Weixiao Huang, Wenhao Wu, Wenyang He, Xianghui Wei, Xianqing Jia, Xingzhe Wu, Xinran Xu, Xinxing Zu, Xinyu Zhou, Xuehai Pan, Y.~Charles, Yang Li, Yangyang Hu, Yangyang Liu, Yanru Chen, Yejie Wang, Yibo Liu, Yidao Qin, Yifeng Liu, Ying Yang, Yiping Bao, Yulun Du, Yuxin Wu, Yuzhi Wang, Zaida Zhou,
  Zhaoji Wang, Zhaowei Li, Zhen Zhu, Zheng Zhang, Zhexu Wang, Zhilin Yang, Zhiqi Huang, Zihao Huang, Ziyao Xu, and Zonghan Yang.
\newblock Kimi k1.5: Scaling reinforcement learning with llms, 2025.

\bibitem{thawakar2025llamav}
Omkar Thawakar, Dinura Dissanayake, Ketan More, Ritesh Thawkar, Ahmed Heakl, Noor Ahsan, Yuhao Li, Mohammed Zumri, Jean Lahoud, Rao~Muhammad Anwer, et~al.
\newblock Llamav-o1: Rethinking step-by-step visual reasoning in llms.
\newblock {\em arXiv preprint arXiv:2501.06186}, 2025.

\bibitem{tong2024cambrian}
Shengbang Tong, Ellis Brown, Penghao Wu, Sanghyun Woo, Manoj Middepogu, Sai~Charitha Akula, Jihan Yang, Shusheng Yang, Adithya Iyer, Xichen Pan, et~al.
\newblock Cambrian-1: A fully open, vision-centric exploration of multimodal llms.
\newblock {\em arXiv preprint arXiv:2406.16860}, 2024.

\bibitem{touvron2023llama}
Hugo Touvron, Thibaut Lavril, Gautier Izacard, Xavier Martinet, Marie-Anne Lachaux, Timoth{\'e}e Lacroix, Baptiste Rozi{\`e}re, Naman Goyal, Eric Hambro, Faisal Azhar, et~al.
\newblock Llama: Open and efficient foundation language models.
\newblock {\em arXiv:2302.13971}, 2023.

\bibitem{wang2023cogvlm}
Weihan Wang, Qingsong Lv, Wenmeng Yu, Wenyi Hong, Ji~Qi, Yan Wang, Junhui Ji, Zhuoyi Yang, Lei Zhao, Xixuan Song, et~al.
\newblock Cogvlm: Visual expert for pretrained language models.
\newblock {\em arXiv:2311.03079}, 2023.

\bibitem{wei2023chainofthoughtpromptingelicitsreasoning}
Jason Wei, Xuezhi Wang, Dale Schuurmans, Maarten Bosma, Brian Ichter, Fei Xia, Ed~Chi, Quoc Le, and Denny Zhou.
\newblock Chain-of-thought prompting elicits reasoning in large language models, 2023.

\bibitem{NEURIPS2022_9d560961}
Jason Wei, Xuezhi Wang, Dale Schuurmans, Maarten Bosma, brian ichter, Fei Xia, Ed~Chi, Quoc~V Le, and Denny Zhou.
\newblock Chain-of-thought prompting elicits reasoning in large language models.
\newblock In S.~Koyejo, S.~Mohamed, A.~Agarwal, D.~Belgrave, K.~Cho, and A.~Oh, editors, {\em Advances in Neural Information Processing Systems}, volume~35, pages 24824--24837. Curran Associates, Inc., 2022.

\bibitem{wen2024efficient}
Yuxin Wen, Qingqing Cao, Qichen Fu, Sachin Mehta, and Mahyar Najibi.
\newblock Efficient vision-language models by summarizing visual tokens into compact registers.
\newblock {\em arXiv preprint arXiv:2410.14072}, 2024.

\bibitem{wu2025mmsearch}
Jinming Wu, Zihao Deng, Wei Li, Yiding Liu, Bo~You, Bo~Li, Zejun Ma, and Ziwei Liu.
\newblock Mmsearch-r1: Incentivizing lmms to search.
\newblock {\em arXiv preprint arXiv:2506.20670}, 2025.

\bibitem{xai2023grok}
{xAI}.
\newblock Grok.
\newblock \url{https://x.ai/}, 2023.
\newblock Large language model.

\bibitem{grok15}
{X.AI}.
\newblock Grok-1.5 vision preview.
\newblock \url{https://x.ai/blog/grok-1.5v}, 2024.

\bibitem{xie2024montecarlotreesearch}
Yuxi Xie, Anirudh Goyal, Wenyue Zheng, Min-Yen Kan, Timothy~P. Lillicrap, Kenji Kawaguchi, and Michael Shieh.
\newblock Monte carlo tree search boosts reasoning via iterative preference learning, 2024.

\bibitem{xing2024pyramiddrop}
Long Xing, Qidong Huang, Xiaoyi Dong, Jiajie Lu, Pan Zhang, Yuhang Zang, Yuhang Cao, Conghui He, Jiaqi Wang, Feng Wu, et~al.
\newblock Pyramiddrop: Accelerating your large vision-language models via pyramid visual redundancy reduction.
\newblock {\em arXiv preprint arXiv:2410.17247}, 2024.

\bibitem{llava-cot}
Guowei Xu, Peng Jin, Li~Hao, Yibing Song, Lichao Sun, and Li~Yuan.
\newblock Llava-o1: Let vision language models reason step-by-step.
\newblock {\em arXiv e-prints}, pages arXiv--2411, 2024.

\bibitem{qwen2.5}
An~Yang, Baosong Yang, Beichen Zhang, Binyuan Hui, Bo~Zheng, Bowen Yu, Chengyuan Li, Dayiheng Liu, Fei Huang, Haoran Wei, Huan Lin, Jian Yang, Jianhong Tu, Jianwei Zhang, Jianxin Yang, Jiaxi Yang, Jingren Zhou, Junyang Lin, Kai Dang, Keming Lu, Keqin Bao, Kexin Yang, Le~Yu, Mei Li, Mingfeng Xue, Pei Zhang, Qin Zhu, Rui Men, Runji Lin, Tianhao Li, Tianyi Tang, Tingyu Xia, Xingzhang Ren, Xuancheng Ren, Yang Fan, Yang Su, Yichang Zhang, Yu~Wan, Yuqiong Liu, Zeyu Cui, Zhenru Zhang, and Zihan Qiu.
\newblock Qwen2.5 technical report.
\newblock {\em arXiv preprint arXiv:2412.15115}, 2024.

\bibitem{yang2024visionzip}
Senqiao Yang, Yukang Chen, Zhuotao Tian, Chengyao Wang, Jingyao Li, Bei Yu, and Jiaya Jia.
\newblock Visionzip: Longer is better but not necessary in vision language models.
\newblock {\em arXiv preprint arXiv:2412.04467}, 2024.

\bibitem{yang2023lidar}
Senqiao Yang, Jiaming Liu, Ray Zhang, Mingjie Pan, Zoey Guo, Xiaoqi Li, Zehui Chen, Peng Gao, Yandong Guo, and Shanghang Zhang.
\newblock Lidar-llm: Exploring the potential of large language models for 3d lidar understanding.
\newblock {\em arXiv preprint arXiv:2312.14074}, 2023.

\bibitem{yang2023improved}
Senqiao Yang, Tianyuan Qu, Xin Lai, Zhuotao Tian, Bohao Peng, Shu Liu, and Jiaya Jia.
\newblock An improved baseline for reasoning segmentation with large language model.
\newblock {\em arXiv preprint arXiv:2312.17240}, 2023.

\bibitem{Yang_2024_CVPR}
Senqiao Yang, Zhuotao Tian, Li~Jiang, and Jiaya Jia.
\newblock Unified language-driven zero-shot domain adaptation.
\newblock In {\em Proceedings of the IEEE/CVF Conference on Computer Vision and Pattern Recognition (CVPR)}, pages 23407--23415, June 2024.

\bibitem{yang2025r1}
Yi~Yang, Xiaoxuan He, Hongkun Pan, Xiyan Jiang, Yan Deng, Xingtao Yang, Haoyu Lu, Dacheng Yin, Fengyun Rao, Minfeng Zhu, et~al.
\newblock R1-onevision: Advancing generalized multimodal reasoning through cross-modal formalization.
\newblock {\em arXiv preprint arXiv:2503.10615}, 2025.

\bibitem{mulberry}
Huanjin Yao, Jiaxing Huang, Wenhao Wu, Jingyi Zhang, Yibo Wang, Shunyu Liu, Yingjie Wang, Yuxin Song, Haocheng Feng, Li~Shen, et~al.
\newblock Mulberry: Empowering mllm with o1-like reasoning and reflection via collective monte carlo tree search.
\newblock {\em arXiv preprint arXiv:2412.18319}, 2024.

\bibitem{yao2024mulberry}
Huanjin Yao, Jiaxing Huang, Wenhao Wu, Jingyi Zhang, Yibo Wang, Shunyu Liu, Yingjie Wang, Yuxin Song, Haocheng Feng, Li~Shen, et~al.
\newblock Mulberry: Empowering mllm with o1-like reasoning and reflection via collective monte carlo tree search.
\newblock {\em arXiv preprint arXiv:2412.18319}, 2024.

\bibitem{yao2023tree}
Shunyu Yao, Dian Yu, Jeffrey Zhao, Izhak Shafran, Tom Griffiths, Yuan Cao, and Karthik Narasimhan.
\newblock Tree of thoughts: Deliberate problem solving with large language models.
\newblock {\em Advances in neural information processing systems}, 36:11809--11822, 2023.

\bibitem{minicpm-v}
Yuan Yao, Tianyu Yu, Ao~Zhang, Chongyi Wang, Junbo Cui, Hongji Zhu, Tianchi Cai, Haoyu Li, Weilin Zhao, Zhihui He, et~al.
\newblock Minicpm-v: A gpt-4v level mllm on your phone.
\newblock {\em arXiv:2408.01800}, 2024.

\bibitem{yu2025dapo}
Qiying Yu, Zheng Zhang, Ruofei Zhu, Yufeng Yuan, Xiaochen Zuo, Yu~Yue, Tiantian Fan, Gaohong Liu, Lingjun Liu, Xin Liu, et~al.
\newblock Dapo: An open-source llm reinforcement learning system at scale.
\newblock {\em arXiv preprint arXiv:2503.14476}, 2025.

\bibitem{yu2024mmvet}
Weihao Yu, Zhengyuan Yang, Linjie Li, Jianfeng Wang, Kevin Lin, Zicheng Liu, Xinchao Wang, and Lijuan Wang.
\newblock Mm-vet: Evaluating large multimodal models for integrated capabilities.
\newblock In {\em ICML}, 2024.

\bibitem{yu2024mm}
Weihao Yu, Zhengyuan Yang, Linjie Li, Jianfeng Wang, Kevin Lin, Zicheng Liu, Xinchao Wang, and Lijuan Wang.
\newblock Mm-vet: Evaluating large multimodal models for integrated capabilities.
\newblock In {\em International conference on machine learning}. PMLR, 2024.

\bibitem{yue2023mmmu}
Xiang Yue, Yuansheng Ni, Kai Zhang, Tianyu Zheng, Ruoqi Liu, Ge~Zhang, Samuel Stevens, Dongfu Jiang, Weiming Ren, Yuxuan Sun, Cong Wei, Botao Yu, Ruibin Yuan, Renliang Sun, Ming Yin, Boyuan Zheng, Zhenzhu Yang, Yibo Liu, Wenhao Huang, Huan Sun, Yu~Su, and Wenhu Chen.
\newblock Mmmu: A massive multi-discipline multimodal understanding and reasoning benchmark for expert agi.
\newblock In {\em Proceedings of CVPR}, 2024.

\bibitem{zhang2024lmms}
Kaichen Zhang, Bo~Li, Peiyuan Zhang, Fanyi Pu, Joshua~Adrian Cahyono, Kairui Hu, Shuai Liu, Yuanhan Zhang, Jingkang Yang, Chunyuan Li, et~al.
\newblock Lmms-eval: Reality check on the evaluation of large multimodal models.
\newblock {\em arXiv preprint arXiv:2407.12772}, 2024.

\bibitem{ixc25}
Pan Zhang, Xiaoyi Dong, Yuhang Zang, Yuhang Cao, Rui Qian, Lin Chen, Qipeng Guo, Haodong Duan, Bin Wang, Linke Ouyang, et~al.
\newblock Internlm-xcomposer-2.5: A versatile large vision language model supporting long-contextual input and output.
\newblock {\em arXiv preprint arXiv:2407.03320}, 2024.

\bibitem{zhang2024mathverse}
Renrui Zhang, Dongzhi Jiang, Yichi Zhang, Haokun Lin, Ziyu Guo, Pengshuo Qiu, Aojun Zhou, Pan Lu, Kai-Wei Chang, Yu~Qiao, et~al.
\newblock Mathverse: Does your multi-modal llm truly see the diagrams in visual math problems?
\newblock In {\em European Conference on Computer Vision}, pages 169--186. Springer, 2024.

\bibitem{llava-reasoner}
Ruohong Zhang, Bowen Zhang, Yanghao Li, Haotian Zhang, Zhiqing Sun, Zhe Gan, Yinfei Yang, Ruoming Pang, and Yiming Yang.
\newblock Improve vision language model chain-of-thought reasoning.
\newblock {\em arXiv preprint arXiv:2410.16198}, 2024.

\bibitem{zhang2024improve}
Ruohong Zhang, Bowen Zhang, Yanghao Li, Haotian Zhang, Zhiqing Sun, Zhe Gan, Yinfei Yang, Ruoming Pang, and Yiming Yang.
\newblock Improve vision language model chain-of-thought reasoning.
\newblock {\em arXiv preprint arXiv:2410.16198}, 2024.

\bibitem{zhang2024sparsevlm}
Yuan Zhang, Chun-Kai Fan, Junpeng Ma, Wenzhao Zheng, Tao Huang, Kuan Cheng, Denis Gudovskiy, Tomoyuki Okuno, Yohei Nakata, Kurt Keutzer, et~al.
\newblock Sparsevlm: Visual token sparsification for efficient vision-language model inference.
\newblock {\em arXiv preprint arXiv:2410.04417}, 2024.

\bibitem{zhong2024lyra}
Zhisheng Zhong, Chengyao Wang, Yuqi Liu, Senqiao Yang, Longxiang Tang, Yuechen Zhang, Jingyao Li, Tianyuan Qu, Yanwei Li, Yukang Chen, et~al.
\newblock Lyra: An efficient and speech-centric framework for omni-cognition.
\newblock {\em arXiv preprint arXiv:2412.09501}, 2024.

\bibitem{zhu2023minigpt}
Deyao Zhu, Jun Chen, Xiaoqian Shen, Xiang Li, and Mohamed Elhoseiny.
\newblock Minigpt-4: Enhancing vision-language understanding with advanced large language models.
\newblock {\em arXiv preprint arXiv:2304.10592}, 2023.

\end{thebibliography}
\bibliographystyle{plain}

\addtocontents{toc}{\protect\setcounter{tocdepth}{2}}

\appendix
\newpage

\setcounter{tocdepth}{-1}  
\hypersetup{linkcolor=cvprblue} 

\tableofcontents
\hypersetup{linkcolor=red}

\section{Related Works}
\label{sec:appendix-related}
\subsection{Efficient Vision Language Models}
Large Language Models (LLMs) have demonstrated remarkable progress in language understanding and generation~\cite{achiam2023gpt, touvron2023llama, bai2023qwen, peng2023instruction, chiang2023vicuna, li2025logits}. Building on their success, VLMs have rapidly advanced by integrating visual information into LLM architectures~\cite{liu2023improvedllava, liu2024llavanext, li2024mini, tong2024cambrian, wang2023cogvlm, liu2024visual, zhu2023minigpt, lai2023lisa, yang2023improved}. Prominent models such as LLaVA~\cite{liu2023improvedllava} utilize visual encoders followed by the projection layers to convert images into token sequences compatible with LLMs. However, as the performance of vision-language models continues to improve, the number of visual tokens grows rapidly, leading to increased computational costs. This trend limits the practical deployment of such models in scenarios like edge computing, autonomous driving, and robotics~\cite{kim24openvla,liu2024robomamba,qu2024mobile,yang2023lidar,Yang_2024_CVPR,minicpm-v,qu2025does}. Therefore, it is imperative to avoid the excessive use of visual tokens.

Recently, some studies~\cite{chen2024image, zhang2024sparsevlm, xing2024pyramiddrop, wen2024efficient, shi2023upop, he2024zipvl, yang2024visionzip,zhong2024lyra} have also recognized the redundancy in visual tokens and proposed various methods to address it. 
Most of these works input images containing redundancy and use the attention scores assigned by the model to prune or merge tokens for token compression.
Furthermore, they typically apply a fixed threshold to compress the same proportion of redundant tokens across all data samples.
Although these methods maintain good performance on general VQA tasks, they perform poorly on OCR-related benchmarks.
In contrast to previous works, our proposed~\methodNAME initially inputs compressed tokens and allows the model to autonomously determine whether token compression is sufficient or if a high-resolution image is required.
Through this approach, our method achieves efficiency while maintaining strong performance on OCR-related benchmarks.
Additionally,~\methodNAME is not a specific token-level compression strategy but represents a new paradigm that can be integrated with existing EfficientVLM methods.

\subsection{Large Language Model Reasoning}
Recent advances in large language models (LLMs)~\cite{NEURIPS2022_9d560961,luong2024reftreasoningreinforcedfinetuning,openai2024openaio1card,kimiteam2025kimik15scalingreinforcement,guo2025deepseek,xai2023grok,team2023gemini,qwen2.5} have significantly improved their reasoning capabilities through methods that simulate human-like stepwise thinking. One foundational technique, Chain-of-Thought (CoT) prompting \cite{wei2023chainofthoughtpromptingelicitsreasoning}, encourages models to decompose complex tasks into intermediate steps, enhancing performance on a variety of reasoning benchmarks.
Furthermore, researchers have explored more structured and dynamic reasoning paradigms, such as Tree-of-Thought and Graph-of-Thought \cite{yao2023tree, besta2024graph}, which organize reasoning as branching or interconnected processes. Complementary approaches like Program-of-Thought (PoT) \cite{chen2023programthoughtspromptingdisentangling} further improve reasoning fidelity by integrating external computational tools to verify or simplify logic steps.

Besides, recent work has also shifted attention from model architecture design and train-time scaling to test-time scaling \cite{snell2024scalingllmtesttimecompute}, such as Monte Carlo Tree Search~(MCTS) \cite{xie2024montecarlotreesearch}, stepwise preference optimization \cite{lai2024stepdpostepwisepreferenceoptimization}, and reinforcement learning \cite{luong2024reftreasoningreinforcedfinetuning} are used to refine outputs during inference. 
Models such as DeepSeek-R1 \cite{guo2025deepseek}, OpenAI-O1~\cite{openai2024openaio1card} demonstrate the effectiveness of combining large-scale RL with reward functions that prioritize both correctness and reasoning quality.
Although LLMs have shown remarkable progress in structured reasoning, extending these abilities to Vision Language Models remains an open challenge.

\subsection{Vision Language Model Reasoning}
With the advancement of LLM reasoning capabilities~\cite{guo2025deepseek}, many studies have aimed to improve the reasoning abilities of VLMs~\cite{zhang2024improve,mitra2024compositional,luan2024textcot}. One common approach is using Chain-of-Thought (CoT) prompting to construct SFT datasets. However, the CoTs generated often lack natural human cognitive processes, limiting their effectiveness and generalization.
Furthermore, inspired by DeepSeek-R1~\cite{guo2025deepseek}, several studies have attempted to transfer this reasoning paradigm to vision tasks~\cite{yao2024mulberry,thawakar2025llamav,huang2025vision,yang2025r1,liu2025seg,shen2025vlm,meng2025mm, liu2025visual, wu2025mmsearch}. Most of these efforts, by collecting CoT data to perform a cold start and then training the model using a reinforcement learning strategy such as GRPO. While this approach achieves performance improvements on specific tasks, it significantly degrades the model's general performance.
Moreover,  current methods remain limited to visual math or segmentation tasks, failing to generalize to broader general VQA tasks.
In this paper, we propose \methodNAME, which effectively applies reinforcement learning to general VQA tasks by leveraging the LLM-as-Judge strategy.


\section{Additional Experiments}
\subsection{Prompt Details}
\label{sec:supp-prompt-details}
\subsubsection{LLM-as-Judge Prompt Design}
In this section, we detail the prompt design for our LLM-as-Judge strategy.
As shown in Table~\ref{table:judge_prompt}, the placeholders 
\textcolor{red}{Ground Truth} and \textcolor{red}{Prediction}  are dynamically replaced with the corresponding question, ground truth answer, and model prediction during evaluation.  
Specifically, the judgment process is conducted entirely in text. Our findings indicate that, compared to VLMs, current LLMs achieve higher judgment accuracy and exhibit fewer hallucinations. Moreover, by eliminating the need for visual token inputs, it significantly reduce the overall evaluation cost.

Furthermore, we require the LLM to return a discrete value, with 1 indicating a correct prediction and 0 indicating an incorrect one, rather than a continuous score representing the degree of correctness. This binary format further reduces the likelihood of misjudgment. In a user study of 1,000 cases, no misclassifications were observed.

\begin{table}[h]
    \centering
    \caption{\textbf{Judgment Prompt Template. } \textcolor{red}{Question}, \textcolor{red}{Ground Truth} and \textcolor{red}{Prediction}  are dynamically replaced with the specific question, ground truth and model prediction during evaluation.}
    
    \begin{tabular}{p{13cm}}
        \toprule
        \textbf{\textit{SYSTEM PROMPT:}} \\
        You are an intelligent chatbot designed for evaluating the correctness of generative outputs for question-answer pairs.\\
        Your task is to compare the predicted answer with the correct answer and determine if they match meaningfully. Here's how you can accomplish the task:\\
        INSTRUCTIONS: \\
        - Focus on the meaningful match between the predicted answer and the correct answer.\\
        - Consider synonyms or paraphrases as valid matches.\\
        - Evaluate the correctness of the prediction compared to the answer. \\
        \midrule
        \textbf{\textit{USER PROMPT:}} \\
        I will give you a question related to an image and the following text as inputs:\\
        1. **Question Related to the Image**: \textcolor{red}{Question}\\
        2. **Ground Truth Answer**: \textcolor{red}{Ground Truth}\\
        3. **Model Predicted Answer**: \textcolor{red}{Prediction}\\
        Your task is to evaluate the model's predicted answer against the ground truth answer, based on the context provided by the question related to the image. Consider the following criteria for evaluation:\\
        - **Relevance**: Does the predicted answer directly address the question posed, considering the information provided by the given question?\\
        - **Accuracy**: Compare the predicted answer to the ground truth answer. You need to evaluate from the following two perspectives:\\
        (1) If the ground truth answer is open-ended, consider whether the prediction accurately reflects the information given in the ground truth without introducing factual inaccuracies. If it does, the prediction should be considered correct.\\
        (2) If the ground truth answer is a definitive answer, strictly compare the model's prediction to the actual answer. Pay attention to unit conversions such as length and angle, etc. As long as the results are consistent, the model's prediction should be deemed correct.\\
        **Output Format**:\\
        Your response should include an integer score indicating the correctness of the prediction: 1 for correct and 0 for incorrect. Note that 1 means the model's prediction strictly aligns with the ground truth, while 0 means it does not.\\
        The format should be Score: 0 or 1\\
        \bottomrule
    \end{tabular}
\label{table:judge_prompt}
\end{table}

\subsubsection{\methodNAME Image Resize Prompt}
\begin{table}[h]
    \centering
    \caption{\textbf{VisionThink Image Resize Prompt Template. } \textcolor{red}{Question} will be replaced with the specific question during training and inference.}
    \begin{tabular}{p{13cm}}
        \toprule
        \textbf{\textit{SYSTEM PROMPT:}} \\
        You are a helpful assistant. \\
        \# Tools \\
        You may call the function tool shown below to assist with the user query. \\
        You are provided with the function signature within <tools></tools> XML tags: \\
        <tools> \\
        \{ \\
        \hspace{10pt} "type": "function",\\
        \hspace{10pt} "function":\{ \\
        \hspace{20pt} "name\_for\_human": "resize\_image", \\
        \hspace{20pt} "name": "resize\_image", \\
        \hspace{20pt} "description": "Resize the image resolution.", \\
        \hspace{30pt} "parameters": \{ \\
        \hspace{30pt} "properties": \{ \\
        \hspace{40pt} "action": \{ \\
        \hspace{50pt} "description": "The action to perform. The available actions are: \\
        \hspace{65pt} \textbf{resize}: Double the resolution of the current image. You should only use this tool if you are unable to obtain the critical information needed to answer the question from the current resolution.", \\
        \hspace{50pt} "enum": ["resize"], \\
        \hspace{50pt} "type": "string" \\
        \hspace{50pt} \} \\ 
        \hspace{40pt} \} \\
        \hspace{30pt} "required": ["action"], \\
        \hspace{30pt} "type": "object", \\
        \hspace{20pt} \}, \\
        \hspace{10pt} "args\_format": "Format the arguments as a JSON object." \\
        \hspace{10pt} \} \\
        \} \\
        </tools> \\
        For each function call, return a json object with the function name and the corresponding argument within <tool\_call></tool\_call> XML tags: \\
        <tool\_call>
        \{"name": <function-name>, "arguments": <args-json-object>\}
        </tool\_call> \\
        \midrule
        \textbf{\textit{USER PROMPT:}} \\
        Answer the question based on the image provided.\ You must conduct reasoning within \textcolor{orange}{\texttt{<think>}} and \textcolor{orange}{\texttt{</think>}} first in each of your reasoning steps.\ You may call ONE function tool per step to help you better solve the problem.\ Place the function tool within \textcolor{cyan}{\texttt{<tool\_call>}} and \textcolor{cyan}{\texttt{</tool\_call>}} at the end of each step to perform a function call.\ You should continue your reasoning process based on the content returned by the function tool.\ Once you confirm your final answer, place the final answer inside \textcolor{purple}{\texttt{<answer>}} and \textcolor{purple}{\texttt{</answer>}}.\ For mathematical or multiple-choice problem, wrap the answer value or choice with \textcolor{darkgreen}{\texttt{\textbackslash boxed\{\}}}. Here is the image and question: \textcolor{red}{Question}.\\
        \bottomrule
    \end{tabular}
\label{table:visionthink_prompt}
\end{table}

As shown in Table~\ref{table:visionthink_prompt}, we present the detailed system and user prompts used in our proposed~\methodNAME. Specifically, we integrate image resizing as a tool-call function. Following the Qwen2.5-VL cookbook~\cite{Qwen2.5-VL}, we employ an Agent Prompt that enables the model to output special tokens to trigger image resizing.
This prompt design allows the model to exhibit distinct behaviors such as requesting image resizing or directly answering the question. These behaviors introduce differentiable gradients, which make it feasible to apply the GRPO algorithm.
Furthermore, we analyze the impact of different prompts in Sec.~\ref{sec:prompt-influence}.

\subsection{Details of the Format Reward}
The format reward has a total score of 0.5, which is awarded only when all formatting requirements are fully satisfied. Specifically, as shown in the \methodNAME Prompt~(Table.~\ref{table:visionthink_prompt}), the first requirement is that the model’s output must include both the the <answer></answer>  and  <think></think> tags, which denote the final answer and the reasoning process, respectively. 
The second requirement states that for responses involving an image resize operation, the model must output a correctly formatted <tool\_call></tool\_call> tag containing a valid JSON content.

\subsection{Implementation Details}
\label{sec:supp-implemantation}
\mypara{Training Details.}
In this paper, we conduct experiments using Qwen2.5-VL-7B-Instruct~\cite{bai2025qwen2.5vl} as the base model, trained with the veRL framework~\cite{sheng2024hybridflow}. We use a total batch size of 512 with mixed-precision (FP16) training. The mini-batch size is set to 32, and the KL divergence coefficient is 0.001. The policy model is optimized using an initial learning rate of $1 \times 10^{-6}$.
For each prompt, we generate 16 candidate responses using a temperature of 1.0, and apply duplicate and empty response filtering, similar to DAPO~\cite{yu2025dapo}. 

\mypara{Inference Details.}
In this paper, we use the lmms-eval~\cite{zhang2024lmms} to evaluate the model's performance. Besides, in order to save the GPU memory and improve the inference speed, we utilize the vLLM\cite{vllm} framework and set the temperature to zero for inference.

\subsection{Benchmark Datasets and Evaluation Metrics}
\label{sec:benchmark}
We conduct experiments on these widely used visual understanding benchmarks.

\mypara{ChartQA. }
ChartQA~\cite{masry2022chartqa} is a benchmark designed to evaluate how well multimodal models answer questions about charts, emphasizing both visual understanding and logical reasoning. It includes various chart types, such as bar charts and line graphs, with a mix of human-written and automatically generated questions to assess complex reasoning abilities.
Notably, ChartQA is a strongly OCR-dependent benchmark that requires fine-grained visual understanding, as models must extract textual information from charts and reason over it.

\begin{figure}
    \centering
    \includegraphics[width=0.95\linewidth]{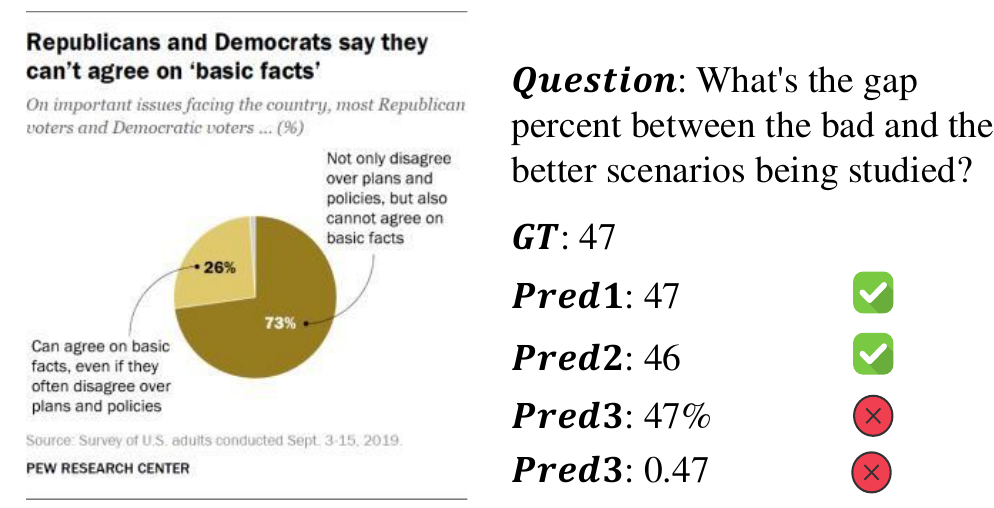}
    \caption{An example illustrating the original evaluation method used in ChartQA.}
    \label{fig:chartqa}
\end{figure}
\lstdefinestyle{mystyle}{
    language=Python,
    backgroundcolor=\color{gray!10},
    commentstyle=\color{green!50!black},
    keywordstyle=\color{blue}\bfseries,
    numberstyle=\tiny\color{gray},
    stringstyle=\color{orange},
    basicstyle=\ttfamily\footnotesize,
    breaklines=true,
    numbers=right,
    numbersep=5pt,
    frame=single,
    rulecolor=\color{gray!70},
    captionpos=b,
    tabsize=4,
    showstringspaces=false
}
\lstset{style=mystyle}

\begin{lstlisting}[caption={Core evaluation code from the original ChartQA assessment method..}, label={lst:ocr}]
def _to_float(text: str):
    try:
        if text.endswith("%"):
            return float(text.rstrip("%")) / 100.0
        else:
            return float(text)
    except ValueError:
        return None

prediction_float = _to_float(prediction)
target_float = _to_float(target)

if prediction_float is not None and target_float is not None:
    relative_change = abs(prediction_float - target_float) / abs(target_float)
    return relative_change <= max_relative_change  # 0.05
else:
    return prediction.lower() == target.lower()
\end{lstlisting}
Furthermore, we observe that the evaluation process of ChartQA in lmms-eval~\cite{zhang2024lmms} relies on a float-value comparison method, which presents several limitations in practical evaluation scenarios.
The corresponding implementation is shown in Listing~\ref{lst:ocr}, and an illustrative example is provided in Fig.~\ref{fig:chartqa} for further analysis.

As shown in Fig.~\ref{fig:chartqa}, for the question ``What’s the gap percent between the bad and the better scenarios being studied?'', the intuitive answer derived from the image is 47\%.
And the \texttt{\_to\_float()} function (Line 1 in Listing~\ref{lst:ocr}) converts both \texttt{0.47} and \texttt{47\%} to \texttt{0.47}, while converting the ground truth value \texttt{47} to \texttt{47.0}.
Hence, the comparison at Line 14 treats both \texttt{0.47} and \texttt{47\%} as incorrect predictions, leading to an erroneous evaluation result.
Moreover, when the model incorrectly predicts \texttt{46}, the current evaluation method still considers it correct, as the relative error compared to the ground truth \texttt{47} is:
$$
\frac{|46 - 47|}{47} = 0.02 < 0.05.
$$
which is also a wrong judgment result of the evaluation. 

Based on this observation, all ChartQA evaluations in this paper are conducted using a combination of GPT-4o-Judge and human verification, denoted as ChartQA$^{\dagger}$.

\mypara{MME.}
The MME benchmark~\cite{fu2023mme} assesses multimodal models on 14 subtasks that reflect both perceptual processing and cognitive reasoning abilities. By utilizing carefully crafted instruction-response pairs, MME aims to minimize the risk of training data contamination, ensuring a fair and rigorous evaluation process.

\mypara{OCRBench.}
OCRBench~\cite{liu2024ocrbench} is a comprehensive benchmark for evaluating the OCR capabilities of vision language models. It covers five key tasks: text recognition, scene text-centric VQA, document-oriented VQA, key information extraction, and handwritten mathematical expression recognition. With 29 datasets and 10,000 human-verified QA pairs across 31 scenarios. Its scenarios span street scenes, receipts, and formulas, testing models on multilingual, handwritten, non-semantic, and mathematical text.

\mypara{DocVQA.}
DocVQA~\cite{docvqa} is a dataset for VQA on document images, comprising 50,000 questions defined on over 12,000 document images. It covers various document types, including forms, receipts, and scientific papers, testing models' ability to understand and reason about document content, such as textual information, tables, and visual elements. 

\mypara{RealWorldQA.}
RealWorldQA~\cite{grok15} is a benchmark designed to evaluate the real-world spatial understanding capabilities of VLMs. It consists of over 700 images, each accompanied by a question and a verifiable answer, drawn from real-world scenarios, including those captured from vehicles. The benchmark assesses how well models comprehend physical environments and spatial relationships, which are crucial for applications in navigation, robotics, and general AI assistance. 

\mypara{MMVet.}
MMVet~\cite{yu2024mm} introduces a structured framework to assess six foundational vision-language skills: recognition, OCR, knowledge, language generation, spatial awareness, and math. These capabilities are combined in 16 evaluation configurations to test how well multimodal systems can integrate them for solving complex tasks, offering a detailed and quantitative performance analysis.

\mypara{POPE.} 
POPE~\cite{li2023evaluating} is designed to measure object hallucination in vision-language models using binary-choice questions that verify whether specific objects are present in given images. It employs metrics such as Accuracy, Recall, Precision, and F1 Score across three distinct sampling strategies, delivering a robust and fine-grained evaluation of hallucination tendencies. In our paper, the result of POPE is F1-score.

\mypara{MMMU.}
MMMU~\cite{yue2023mmmu} serves as a benchmark for assessing multimodal models on intricate, college-level tasks that demand both extensive knowledge and reasoning capabilities. It comprises  11.5K carefully selected questions sourced from exams, quizzes, and academic textbooks, spanning six broad fields: Art \& Design, Business, Science, Health \& Medicine, Humanities \& Social Sciences, and Technology \& Engineering. These questions encompass 30 academic subjects and 183 specialized areas, incorporating a wide variety of visual formats such as diagrams, graphs, and chemical formulas. MMMU is designed to push models toward expert-level performance by testing their ability to understand and reason across disciplines and modalities.

\mypara{MathVista.}
MathVista~\cite{mathvista} is a benchmark for evaluating the mathematical reasoning capabilities of foundation models within visual contexts. It includes 6,141 examples, derived from 28 existing multimodal datasets involving mathematics and three newly created datasets: IQTest, FunctionQA, and PaperQA. These tasks require fine-grained visual understanding and compositional reasoning, often involving the interpretation of graphs, equations, and other mathematical visuals. MathVista aims to systematically study the ability of VLMs to solve mathematical problems presented in visual formats, highlighting the need for models that can seamlessly integrate visual perception with mathematical reasoning.

\mypara{MathVerse.}
MathVerse~\cite{zhang2024mathverse} is a benchmark for rigorously evaluating the capabilities of VLMs in interpreting and reasoning with visual information in mathematical problems. MathVerse consists of 2,612 high-quality, multi-subject math problems with diagrams, each transformed into six distinct versions with varying degrees of information content in multi-modality, resulting in 15,000 test samples.

\subsection{Scaling-up Reinforcement Learning on General VQA Tasks}
\label{sec:supp-scaleup}
Due to the diversity and complexity inherent in general VQA tasks, traditional rule-based reinforcement learning algorithms are not directly applicable. To overcome this limitation, we introduce an LLM-as-Judge strategy, which enables our model to be trained via reinforcement learning on the General VQA task. 
To further demonstrate the effectiveness of our method, we scale up the dataset size to 130K to validate its effectiveness.

\mypara{Dataset.}
Since the LLM-as-Judge approach is flexible, one advantage is that most of the SFT data can be utilized. Therefore, we only filter out subjective open-ended questions whose answers are not unique and can be correctly addressed from different perspectives, such as image descriptions, essay writing, and similar tasks.
Based on this, we ultimately filtered 130K QA pairs to train the VLM via reinforcement learning, without requiring any cold-start phase. All the data will be open-sourced.

\mypara{Prompt.}
To verify the effectiveness of the LLM-as-Judge strategy on general VQA tasks, we conduct experiments with minimal modifications to both the system and user prompts. The detailed prompts are provided in Table~\ref{table:generalqa_prompt}.

\mypara{Reward.}
Since the entire training process in this setting does not involve any decision-making regarding the need for high-resolution images, the total reward function in reinforcement learning is designed to focus solely on answer quality and response formatting. Specifically, the reward comprises two components:
The first component is an accuracy reward, evaluated by the LLM-as-Judge. This component assesses the correctness of the model’s answer against the ground truth, with a maximum of 1 point awarded for a fully correct response.
The second component is a formatting reward, worth 0.5 points. This is granted when the model correctly wraps its response using both the <answer></answer> and <think></think> tags. These tags are critical for maintaining consistent output formatting and enabling downstream interpretability.

\begin{table}[h]
    \centering
    \caption{\textbf{Prompt Template for VisionThink$^\spadesuit$.} VisionThink$^\spadesuit$ refers to a model trained on general VQA tasks using full image resolution and the LLM-as-Judge strategy. The \textcolor{red}{Question} placeholder is replaced with the actual question during training and inference.}
    \begin{tabular}{p{13cm}}
        \toprule
        \textbf{\textit{SYSTEM PROMPT:}} \\
        You FIRST think about the reasoning process as an internal monologue and then provide the final answer. \\
        The reasoning process MUST BE enclosed within \textcolor{orange}{\texttt{<think>}} \textcolor{orange}{\texttt{</think>}} tags. The final answer MUST BE put within \textcolor{purple}{\texttt{<answer>}} \textcolor{purple}{\texttt{</answer>}} tags. For mathematical or multiple-choice problems, wrap the answer value or choice with \textcolor{darkgreen}{\texttt{\textbackslash boxed\{\}}}.\\
        \midrule
        \textbf{\textit{USER PROMPT:}} \\
        \textcolor{red}{Question}.\\
        \bottomrule
    \end{tabular}
\label{table:generalqa_prompt}
\end{table}

\mypara{Experimental Results.}
As shown in Table~\ref{table:generalqa_result}, we compare our model against state-of-the-art open-source and closed-source vision-language models across several general VQA benchmarks. In this evaluation, \methodNAME$^\spadesuit$ denotes our model variant trained using the proposed LLM-as-Judge strategy with the above reward function and 130K QA pairs.

The experimental results demonstrate that our method outperforms the baseline model, Qwen2.5VL-7B-Instruct, across multiple benchmarks. The improvement is particularly notable on the MMVet, where our model achieves a significant performance gain of 7.9\% over the baseline. This highlights the model’s superior capability in handling general VQA tasks.
Furthermore, on the recently popular benchmark MathVista ~\cite{mathvista}, which is designed to assess both mathematical reasoning and general visual question answering reasoning abilities, our model achieves a score of 71.2. This result not only surpasses all existing open-source models but also outperforms several closed-source models. These findings provide strong empirical evidence for the effectiveness and generalizability of our LLM-as-Judge strategy in enhancing the reasoning capabilities of VLMs across general VQA tasks.

\begin{table*}[t]
  \centering
  \caption{\textbf{Effectiveness of LLM-as-Judge Accuracy Reward Design.} VisionThink$^\spadesuit$ is a model we developed by training with the LLM-as-Judge on 130K filtered General VQA datasets and leverages Qwen2.5-VL-7B-Instruct as the base model. Qwen2.5-VL-7B$^{*}$ reports the results evaluate by lmms-eval\cite{zhang2024lmms}.}
  \begin{adjustbox}{width=\textwidth}
  \renewcommand{\arraystretch}{1.5}
  \setlength{\tabcolsep}{2pt}
  \begin{tabular}{@{}l|ccccccccc@{}}
    \toprule
    \multirow{2}{*}{Method} & \textbf{MMMU} & \textbf{MMMU-Pro} & \textbf{MMBench} &\textbf{RealWorldQA} & \textbf{POPE} & \textbf{MME} & \textbf{MathVista} & \textbf{MathVerse} & \textbf{MMVet} \\ 
    \cline{2-10}
     & val & test & en\_test & test & test & test & testmini & testmini & test \\
    \midrule
    \multicolumn{9}{l}{\textit{Closed-Source Model}} \\
    GPT-4o~\cite{gpt4o} & 69.1 & 54.0 & 83.4 & 58.6 & 85.6 & 2329 & 63.8 & 50.2 & 69.1 \\
    Claude-3.5 Sonnet~\cite{sonnet3_5} & 68.3 & 55.0 & 82.6 & 59.9 & - & 1920 & 67.7 & 41.2 & 70.1 \\
    Gemini-1.5-Pro~\cite{team2023gemini} & 62.2 & 49.4 & 73.9 & 70.4 & 88.2 & - & 63.9 & - & 64.0 \\
    \midrule
    \multicolumn{10}{l}{\textit{Open-Source General Model}} \\ 
    Cambrain-1-8B~\cite{tong2024cambrian} & 42.7 & - & 75.9 & 60.0 & 86.4 & 1803 & 49.0 & - & - \\  
    InternVL2-8B~\cite{internvl2} & 49.3 & 32.5 & 81.7 & 64.4 & 84.2 & 2210 & 58.3 & - & 60.0 \\
    LLaVA-OneVision-7B~\cite{llavaov} & 48.8 & - & - & 66.3 & 88.4 & 1998 & 63.2 & - & 57.5 \\
    MiniCPM-Llama-V-2.5-8B~\cite{minicpm-v} & 45.8 & 19.6 & 77.2 & 63.0 & 86.7 & 2025 & 54.3 & - & - \\
    MiniCPM-V-2.6-8B~\cite{minicpm-v} & 49.8 & 27.2 & 78.0 & 65.0 & 83.2 & 2348 & 60.6 & - & - \\
    IXC-2.5~\cite{ixc25} & 42.9 & - & 82.2 & 67.8 & - & 2229 & 63.8 & - & 51.7 \\
    InternVL2.5-8B~\cite{internvl25} & 56.0 & 38.2 & 84.6 & 70.1 & 90.6 & 2344 & 64.4 & 39.5 & 62.8 \\
    \midrule
    \multicolumn{10}{l}{\textit{Reasoning Model}} \\ 
    LLaVA-CoT-11B~\cite{llava-cot} & - & - & 75.0 & - & - & - & 54.8 & - & 60.3 \\
    LLaVA-Reasoner-8B~\cite{llava-reasoner} & - & - & - & - & - & - & 50.6 & - & - \\
    Insight-V-8B~\cite{insightv} & 50.2 & 24.9 & 82.3 & - & - & 2312 & 59.9 & - & - \\
    Mulberry-7B~\cite{mulberry} & 55.0 & - & - & - & - & 2396 & 63.1 & - & - \\
    Vision-R1-LlamaV-CI-11B~\cite{visionr1} & - & - & - & - & - & 2190 & 62.7 & 27.1 & - \\
    \midrule
    \rowcolor{mygray}
    \multicolumn{10}{l}{\textit{\methodNAME}} \\
    Qwen2.5-VL-7B$^{*}$~\cite{bai2025qwen2.5vl} & 50.3 & 37.7 & 82.6 & 68.6 & 86.7 & 2316 & 68.2 & 46.3 & 61.6 \\
    \methodNAME$^\spadesuit$ & \textbf{52.7} & \textbf{41.1} & \textbf{83.4} & 66.5 & \textbf{88.6} & 2314 & \textbf{71.2} & \textbf{48.3} & \textbf{69.5} \\
    \bottomrule
  \end{tabular}
  \end{adjustbox}
  \label{table:generalqa_result}
\end{table*}

\subsection{Comparison with Previous Efficient VLM}
\renewcommand{\multirowsetup}{\centering}
\definecolor{mygray}{gray}{.92}
\definecolor{ForestGreen}{RGB}{34,139,34}
\definecolor{Forestred}{RGB}{220,50,50}
\begin{table*}[t]
    \centering
	\caption{\textbf{Comparison with Previous Efficient VLM Methods.} Vanilla represents the Qwen2.5-VL-7B-Instrcut. The retained ratio of the baseline methods is a predefined hyperparameter, while for \methodNAME, the ratio is determined autonomously by the model and reported as a statistical value. Note that \textit{Down-Sample} refers to the model's performance when directly fed images with their resolution reduced by half. VisionZip\ddag \space represents using the 130K data to finetuning the model.}
    \vspace{0.1cm}
    \begin{adjustbox}{width=\textwidth}
    \setlength{\tabcolsep}{2pt}
    \renewcommand{\arraystretch}{1.6}
    \footnotesize
	\centering
    \begin{tabular}{p{3.5cm}|c c c c c c c c c| c}
        \toprule
        \multirow{2}{*}{\textbf{Method}} & \textbf{ChartQA$^{\dagger}$} & \textbf{OCRBench} & \textbf{DocVQA} & \textbf{MME} & \textbf{MMVet} & \textbf{RealWorldQA} & \textbf{POPE} & \textbf{MathVista} & \textbf{MathVerse} & \multirow{2}{*}{\makecell[c]{\textbf{Avg}.}}\\
        \cline{2-10}
        & test & test & val & test & test & test & test & testmini & testmini &  \\
        \midrule
        \rowcolor{mygray}
        \multicolumn{11}{c}{\textit{Retain 100\% Visual Tokens Across All Benchmarks}}\\
        \multirow{2}{*}{Vanilla} & 79.8 & 81.5 & 95.1 & 2316 & 61.6 & 68.6 & 86.7 & 68.2 & 46.3 & \multirow{2}*{100\%} \\
        ~ & 100\% & 100\% & 100\% & 100\% & 100\% & 100\% & 100\% & 100\% & 100\%  & ~ \\
        \hline
        \rowcolor{mygray}
        \multicolumn{11}{c}{\textit{Retain 25\% Visual Tokens Across All Benchmarks}}\\
        \multirow{2}{*}{Down-Sample} & 62.9 & 68.8 & 94.3 & 2270 & 54.5 & 68.8 & 82.8 & 62.2 & 43.1 & \multirow{2}*{92.1\%} \\
        ~ & 78.8\% & 84.4\% & 99.1\% & 98.0\% & 88.5\% & 100.3\% & 95.5\% & 91.2\% & 93.1\%  & ~ \\
        \hline
        \rowcolor{mygray}
        \multicolumn{11}{c}{\textit{Retain 50\% Visual Tokens Across All Benchmarks}} \\
        \multirow{2}{*}{FastV~(ECCV 2024)} & 72.6 & 75.8 & 93.6 & 2308 & 52.8 & 68.8 & 84.7 & 63.7 & 45.0 & \multirow{2}*{\textbf{95.8\%}} \\
        ~ & 91.0\% & 93.0\% & 98.4\% & 99.6\% & 85.7\% & 100.3\% & 97.7\% & 93.4\% & 97.2\% & ~\\
        \hline
        \multirow{2}{*}{SparseVLM~(ICML 2025)} & 73.2 & 75.6 & 66.8 & 2282 & 51.5  & 68.4 & 85.5 & 66.6 & 45.1 & \multirow{2}*{\textbf{92.2\%}} \\
        ~ & 91.7\% & 92.7\% & 70.2\% & 98.5\% & 83.6\% & 99.7\% & 98.6\% & 97.6\% & 97.4\% &  ~ \\
        \hline
        \multirow{2}{*}{VisionZip~(CVPR 2025)} & 73.4 & 70.5 & 93.8 & 2209 & 57.0 & 68.6 & 86.3 & 63.7 & 45.1 & \multirow{2}*{\textbf{95.0\%}} \\
        ~ & 92.0\% & 86.5\% & 98.6\% & 95.4\% & 92.5\% & 100\% & 99.5\% & 93.4\% & 97.4\% & ~\\
        \hline
        \multirow{2}{*}{VisionZip\ddag~(CVPR 2025)} & 77.3 & 77.9 & 93.8 & 2244 &50.1 & 69.2 & 91.2 & 63.1 & 39.4 & \multirow{2}*{\textbf{95.0\%}} \\
        ~ & 96.9\% & 95.6\% & 98.6\% & 96.9\% & 81.3\% & 100.9\% & 107.5\% & 92.5\% & 85.1\% & ~\\
        \hline
        \rowcolor{mygray}
        \multicolumn{11}{c}{\textit{Retain 70\% Visual Tokens Across All Benchmarks}} \\
        \multirow{2}{*}{FastV~(ECCV 2024)} & 77.2 & 82.2 & 94.4 & 2342 & 56.0 & 68.6 & 85.9 & 65.9 & 46.9 & \multirow{2}*{\textbf{98.4\%}} \\
        ~ & 96.7\% & 100.8\% & 99.3\% & 101.1\% & 90.9\% & 100.0\% & 99.1\% & 96.6\% & 101.3\% & ~\\
        \hline
        \multirow{2}{*}{SparseVLM~(ICML 2025)} & 75.8 & 79.3 & 68.7 & 2276 & 53.7 & 68.5 & 85.4 & 66.3 & 45.1 & \multirow{2}*{\textbf{93.6\%}} \\
        ~ & 94.9\% & 97.3\% & 72.2\% & 98.3\% & 87.2\% & 99.8\% & 98.5\% & 97.2\% & 97.4\% &  ~ \\
        \hline
        \multirow{2}{*}{VisionZip~(CVPR 2025)} & 76.8 & 80.9 & 94.5 & 2334 & 60.0 & 68.2 & 86.4 & 68.9 & 45.8 & \multirow{2}*{\textbf{99.1\%}} \\
        ~ & 96.2\% & 99.3\% & 99.4\% & 100.8\% & 97.4\% & 99.4\% & 99.7\% & 101.0\% & 98.9\% & ~\\
        \hline
        \multirow{2}{*}{VisionZip\ddag~(CVPR 2025)} & 78.2 & 81.3 & 94.1& 2230 & 52.5 & 68.6 & 92.5 & 64.8& 41.8 & \multirow{2}*{\textbf{96.7\%}} \\
        ~ & 98.0\% & 99.8\% & 98.9\% & 96.3\% & 85.3\% & 100\% & 106.7\% & 95.0\% & 90.3\% & ~\\
        \hline
        \rowcolor{mygray}
        \multicolumn{11}{c}{\textit{Retain Approximately 51.3\% Visual Tokens Across All Benchmarks}} \\
        \multirow{2}{*}{\methodNAME} & 79.8 & 80.8 & 94.4 & 2400 & 68.5 & 67.1 & 86.0 & 67.5 & 48.0 & \multirow{2}*{$\fr{101.4\%}$} \\
        ~ & 100.0\% & 99.1\% & 99.3\% & 103.6\% & 111.2\% & 97.8\% & 99.2\% & 99.0\% & 103.7\% &  ~ \\
        \bottomrule
	\end{tabular}
      \end{adjustbox}

	\label{table:efficiency_supp}
\end{table*}
To further demonstrate the effectiveness of our proposed \methodNAME, we conduct a comparative analysis against an additional efficient Vision-Language Model (VLM), VisionZip~\cite{yang2024visionzip}. While previous methods such as FastV~\cite{chen2024image} and SparseVLM~\cite{zhang2024sparsevlm} perform token compression within the language model component based on attention scores, VisionZip applies compression directly within the vision encoder using a similar attention-based mechanism. 

As shown in Table~\ref{table:efficiency_supp}, although previous efficient VLM methods achieve competitive performance on general VQA benchmarks, their accuracy drops significantly on OCR-related tasks. 
This degradation is particularly evident even when a substantial portion of the visual token is retained~(70\%), as demonstrated on the ChartQA dataset.

In contrast, our proposed model, \methodNAME, can smartly decide whether to request the original high-resolution image based on the complexity and demands of each sample. This adaptive strategy enables the model to maintain high accuracy on general VQA tasks while substantially improving performance on benchmarks requiring detailed textual recognition.
Through this capability, \methodNAME demonstrates stronger fine-grained visual understanding and addresses a key limitation of previous efficient VLMs—namely, their poor performance on OCR-related tasks, which has constrained their applicability in real-world scenarios.

Notably, VisionZip$\ddag$ refers to the variant fine-tuned on the 130K dataset~\cite{yang2024visionzip}. However, compared to the training-free version of VisionZip, this fine-tuned model does not show any performance improvement. We attribute this to the limited coverage and diversity of the fine-tuning dataset, which falls short of the supervised fine-tuning data used by the official Qwen team.
This observation indirectly suggests that, compared to supervised fine-tuning, reinforcement learning provides better generalization, which we further discuss in Sec.~\ref{sec:notsft}.

\section{Further Discussions}
\label{sec:supp-discuss}
\subsection{Why Use RL Instead of SFT?}
\label{sec:notsft}
In this paper, we train a smart and efficient vision-language model via reinforcement learning. A natural question arises: why use reinforcement learning instead of supervised fine-tuning to achieve this goal?

To answer this question, we conduct a comparative SFT experiment.
Firstly, we construct the SFT training set. Specifically, compared to RL, which can directly utilize QA pairs and autonomously learn both the reasoning process and whether a high-resolution image is needed, SFT requires manually crafting both the reasoning steps and dialogue that involves high-resolution image requests.
To overcome this limitation, we use GPT-4o to simulate both the high-resolution image requests and the corresponding reasoning process, enabling the SFT training data to closely approximate the behavior of the RL-trained model. Finally, we convert the original RL training data into a format compatible with SFT, maintaining a 1:1 ratio between high-resolution requests and direct answers.

As shown in Table~\ref{table:rl_sft_ratio}, we compare the proportion of high-resolution image requests made by the SFT and RL models across evaluation benchmarks.
Compared to the RL-trained model, which can smartly decide when to answer directly and when to request a high-resolution image, the SFT model exhibits a significantly higher image resize calling ratio across all benchmarks.
This behavior is especially evident in the RealWorldQA benchmark, where high-resolution images are generally unnecessary, yet the SFT model still issues requests 62.1\% of the time.

Based on this observation, we find that SFT does not enable the model to become ``smart'' enough to accurately determine whether a high-resolution image is necessary and also requires constructing the training set using GPT-4o. In contrast, RL makes the VLM smarter and more generalizable, and can directly use the original QA pairs without additional formatting.

\renewcommand{\multirowsetup}{\centering}
\definecolor{mygray}{gray}{.92}
\definecolor{ForestGreen}{RGB}{34,139,34}
\definecolor{Forestred}{RGB}{220,50,50}
\begin{table*}[t]
\centering
\caption{Comparison of image resize call ratios for RL and SFT trained models over multiple evaluation benchmarks.}
    \vspace{0.1cm}
    \setlength{\tabcolsep}{2pt}
    \renewcommand{\arraystretch}{1.6}
    \footnotesize
    \centering
    \begin{tabular}{p{1.2cm}|c c c c c c}
    \toprule
    \textbf{Method} & \textbf{ChartQA$^{\dagger}$} & \textbf{OCRBench} & \textbf{DocVQA} & \textbf{MME} & \textbf{RealWorldQA} & \textbf{POPE} \\
    \midrule
    RL & 79.1\% & 62.3\% & 6.5\% & 30.7\% & 29.9\% & 9.5\% \\
    \hline
    SFT & 95.1\% & 64.0\% & 14.1\% & 39.0\% & 62.1\% & 37.8\% \\
    \bottomrule
    \end{tabular}
\label{table:rl_sft_ratio}
\end{table*}

\subsection{Why not Cold-Start?}
\label{sec:supp-no-cold}
Currently, most explorations of RL in VLMs require a cold start stage. In this section, we explore why we do not use a cold start and instead train the model directly with RL.

To investigate this problem, we collect datasets of 2K and 8K samples to cold start our model. The cold-start data are constructed similarly to Sec.~\ref{sec:notsft}, where GPT-4o is used to simulate both the requests for high-resolution images and the corresponding reasoning processes, enabling the SFT training data to closely mimic the behavior of the model trained via RL.

We first compare the performance of the cold-started models before RL training, as shown in the `Base' lines of Table~\ref{table:coldstart_rl}. Although performance improves with increasing data size, the cold-start models still fall short of the original Qwen2.5VL. We believe this is primarily due to the limited diversity and coverage of our data compared to the SFT data used by the Qwen team, resulting in the observed performance gap.

Furthermore, we compare the performance of models after RL training, using the same RL setup as \methodNAME, as shown in the `RL' lines of Table~\ref{table:coldstart_rl}. 
While RL improves the performance of cold-start models, they still fall short compared to models trained from vanilla Qwen2.5VL with RL.

Based on this observation, we conclude that due to the lower diversity and coverage of the cold-start data compared to the original Qwen2.5VL SFT data, introducing cold-start training may improve performance in specific domains covered by the cold-start data but significantly reduces the model’s general capability. This limitation restricts its broader applicability. Therefore, in this paper, we do not utilize the cold-start stage.

\renewcommand{\multirowsetup}{\centering}
\definecolor{mygray}{gray}{.92}
\definecolor{ForestGreen}{RGB}{34,139,34}
\definecolor{Forestred}{RGB}{220,50,50}
\begin{table*}[t]
    \centering
	\caption{\textbf{Performance comparison of the cold start model and no cold start model.} The without cold start model represents our \methodNAME.}
    \vspace{0.1cm}
    \setlength{\tabcolsep}{2pt}
    \renewcommand{\arraystretch}{1.6}
    \footnotesize
	\centering
    \begin{tabular}{p{2.3cm}|c| c c c c c c c}
    \toprule
    \textbf{Method} & \textbf{Type} & \textbf{ChartQA$^{\dagger}$} & \textbf{OCRBench} & \textbf{DocVQA} & \textbf{MME} & \textbf{MMVet} & \textbf{RealWorldQA} & \textbf{POPE} \\
    \midrule
    w/o Cold Start & RL & 79.8 & 80.8 & 94.4 & 693/1707 & 68.5 & 67.1 & 86.0 \\
    \midrule
    \multirow{2}{*}{Cold Start (2K)} & Base & 76.4 & 78.7 & 92.4 & 444/1354 & 58.3 & 47.2 & 86.6 \\
    & RL & 77.7 & 80.2 & 93.0 & 622/1624 & 62.3 & 52.8 & 86.2 \\
    \midrule
    \multirow{2}{*}{Cold Start (8K)} & Base & 76.8 & 78.2 & 90.5 & 525/1368 & 60.5 & 36.5 & 84.8 \\
    & RL & 79.2 & 79.4 & 92.5 & 609/1637 & 66.2 & 55.8 & 85.6 \\
    \bottomrule
\end{tabular}
\label{table:coldstart_rl}
\end{table*}

\subsection{Different Prompt Impact}
\label{sec:prompt-influence}
Since we do not adopt a cold-start stage, designing an appropriate initial prompt becomes crucial to ensure the model begins from a good starting point.
The base model Qwen2.5VL-Instruct, which has not undergone RL training, typically tends to answer questions directly and lacks the ability to smartly request high-resolution images when needed.

\begin{figure}[h]
    \centering
    \includegraphics[width=0.95\linewidth]{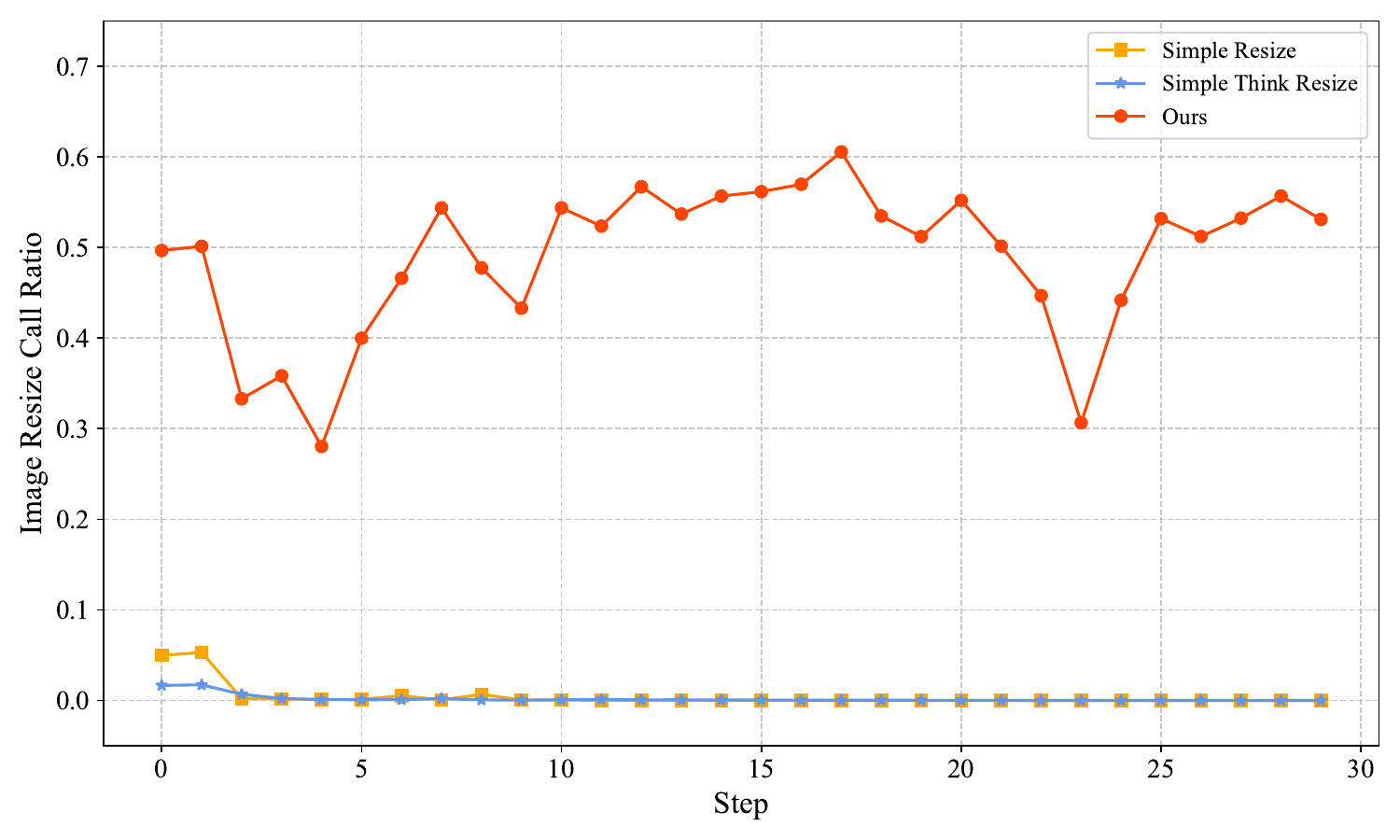}
    \caption{\textbf{Impact of Prompt Choice.} Prompts lead to substantial variation in image resize call ratios, with the Qwen official agent prompt demonstrating the most effective performance.}
    \label{fig:ablation-prompt}
\end{figure}

Therefore, it is essential for the base model, when conditioned on our prompt, to show some preference for calling image resizing. Otherwise, if it is overly biased toward direct answering, the GRPO training process will fail to optimize effectively and may collapse into the direct-answering mode.
To address this, we compare three prompt settings. The first is the official agent prompt from Qwen's cookbook, shown in Table~\ref{table:visionthink_prompt}. The other two are our custom-designed prompts, detailed in Table~\ref{table:custom_prompt}.

\begin{table}[h]
    \centering
    \caption{\textbf{Two custom prompts for analyzing the impact of different prompts.} The \textcolor{red}{Question} placeholder will be replaced with the specific question during training and inference.}
    \begin{tabular}{p{13cm}}
        \toprule
        \textbf{\textit{Simple Resize System Prompt:}} \\
        You are a helpful assistant. \\
        \hline
        \textbf{\textit{Simple Resize User Prompt:}} \\
        Answer the user's question based on the image provided. You can place \textcolor{cyan}{\texttt{<resize></resize>}} at the end of your response to call the image resize tool, it will return the resized image with its resolution doubled to help you better answer the question. Once you confirm your final answer, place the final answer inside \textcolor{purple}{\texttt{<answer>}} and \textcolor{purple}{\texttt{</answer>}}. \\
        Here is the image and question: \textcolor{red}{Question}.\\
        \midrule
        \textbf{\textit{Simple Think Resize System Prompt:}} \\
        You are a helpful assistant. Answer the user's question based on the image provided. You can place \textcolor{cyan}{\texttt{<resize>}} at the end of your response to call the image resize tool, it will return the resized image with its resolution doubled to help you better answer the question. Once you confirm your final answer, place the final answer inside \textcolor{purple}{\texttt{<answer>}} and \textcolor{purple}{\texttt{</answer>}}. \\
        \hline
        \textbf{\textit{Simple Think Resize User Prompt:}} \\
        Enclose reasoning in \textcolor{orange}{\texttt{<think>}}\textcolor{orange}{\texttt{</think>}}\\
        \textcolor{red}{Question}.\\
        \bottomrule
    \end{tabular}
\label{table:custom_prompt}
\end{table}

As shown in Fig.~\ref{fig:ablation-prompt}, our model using the official agent prompt demonstrates a higher image resize call ratio on the effective batch, which consists of an equal mix of high-resolution-required samples and direct-answer samples. In contrast, the other two custom-designed prompts perform poorly, both in their initial behavior and in the final trained model's ability to correctly request image resizing, and ultimately collapse into consistently producing direct answers. Based on these analyses, we find that the Qwen official agent prompt, likely optimized during the pretraining or supervised fine-tuning stages by the Qwen team, is more suitable for \methodNAME.

\subsection{Ablation Study on Penalty Control Threshold}
\begin{figure}
    \centering
    \includegraphics[width=0.95\linewidth]{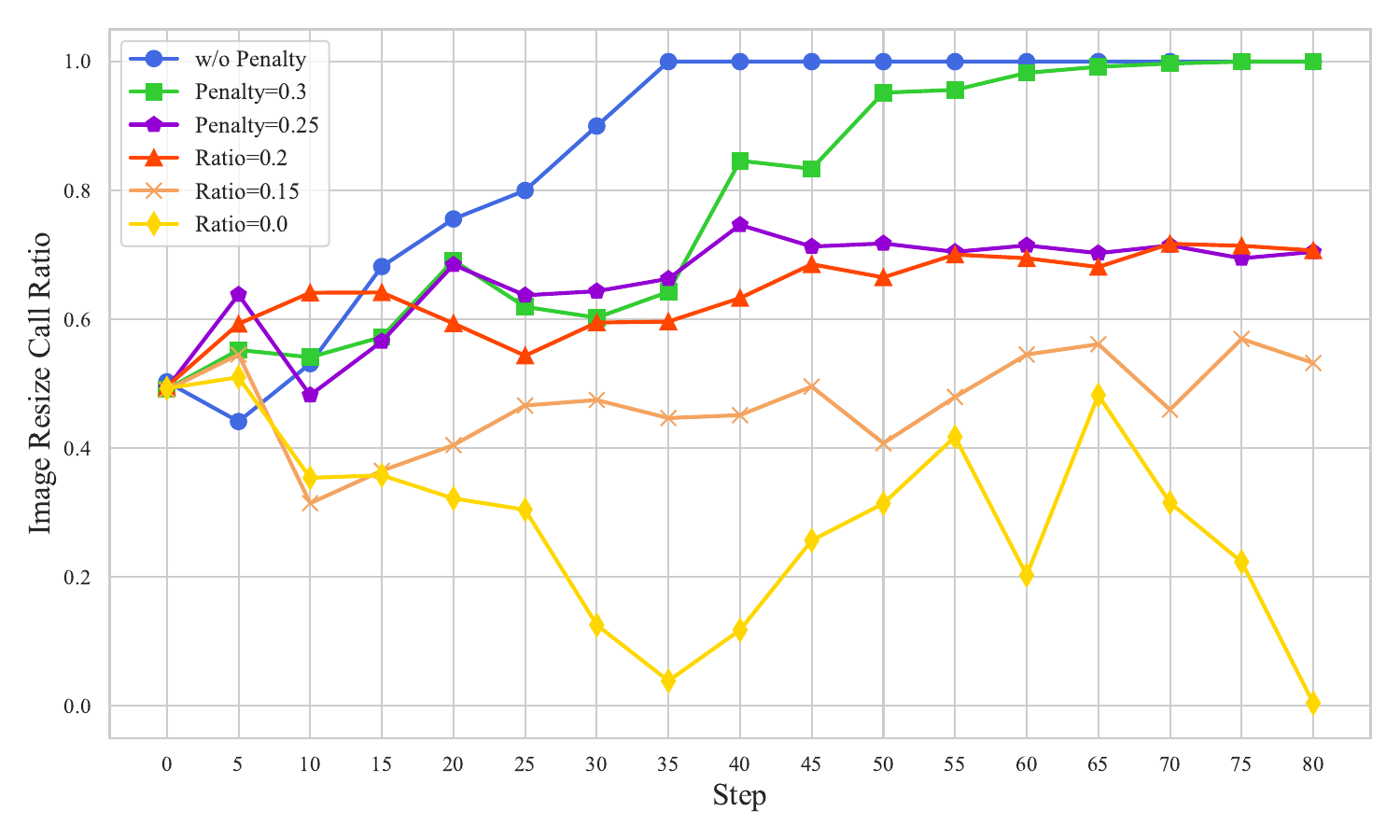}
    \caption{\textbf{Ablation Study on Penalty Ratio Threshold.} As the threshold increases, the model progressively favors requesting image resizing instead of providing direct answers.}
    \label{fig:ablation-penalty}
\end{figure}

In the main paper Eq.~\ref{eq:penalty}, we design the penalty ratio as below:
\begin{equation}
\mathcal{P}_{control}=0.1\cdot\left[\mathbf 1_{\text{direct}}\mathbb{I}(r<\theta)+\mathbf 1_{\text{high}}\mathbb{I}(r\ge\theta)\right],\qquad
r=\frac{C_{\text{direct}}}{C_{\text{direct}}+C_{\text{high}}},
\label{eq:penalty-sup}
\end{equation}
where $\theta$ is the threshold.

Intuitively, the larger the value of $\theta$, the more likely the model is to penalize direct answers, thereby encouraging it to request high-resolution images. Conversely, a smaller $\theta$ leads the model to penalize responses that call for image resizing, thus promoting direct answers.
Based on this intuition, we experimented with different threshold values and recorded the proportion of high-resolution image requests within the effective batch. The results are shown in Fig.~\ref{fig:ablation-penalty}. As indicated by the Eq.~\ref{eq:penalty-sup}, increasing the threshold gradually shifts the model's behavior from favoring direct answers to favoring image resizing requests. Eventually, the model collapses into always requesting high-resolution images. However, within an appropriate range, the model’s behavior is not highly sensitive to the exact threshold value.

\section{Qualitative Results}
\label{sec:supp-qualitative}
In this section, we present a case study comparing our proposed~\methodNAME with other efficient VLM methods: FastV~\cite{chen2024image}, SparseVLM~\cite{zhang2024sparsevlm}, and VisionZip~\cite{yang2024visionzip}.
As shown in the three cases below, for OCR-related or detail-intensive samples, our proposed \methodNAME model can smartly determine when a high-resolution image is needed.
In contrast to previous efficient VLMs, which suffer performance degradation due to fixed compression ratios, \methodNAME avoids such issues by making adaptive decisions based on the input, thereby maintaining strong performance.

\begin{figure}[h]
    \centering
    \includegraphics[width=\linewidth]{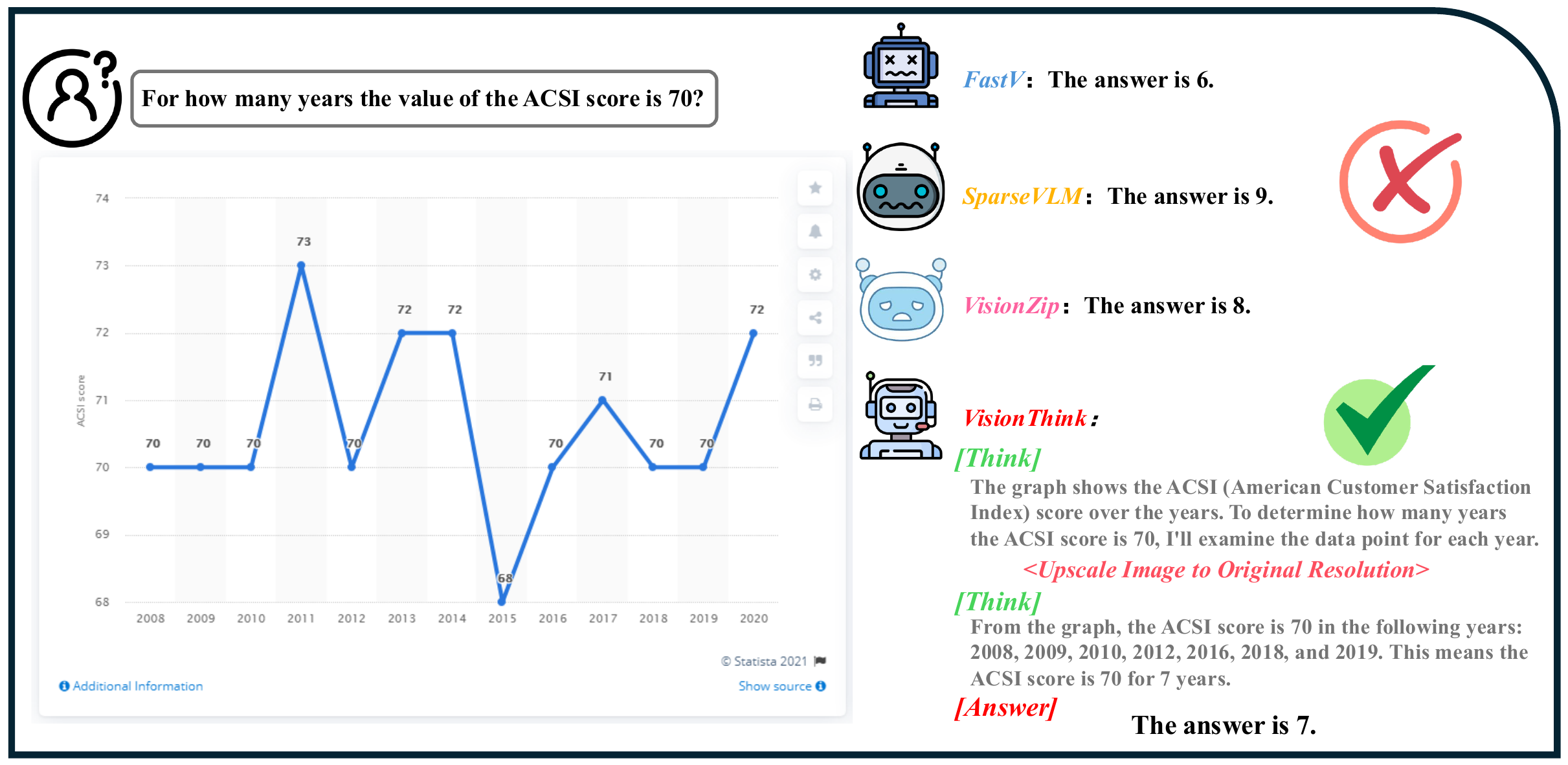}
    \label{fig:sup-case2}

\end{figure}

\begin{figure}[h]
    \centering
    \includegraphics[width=\linewidth]{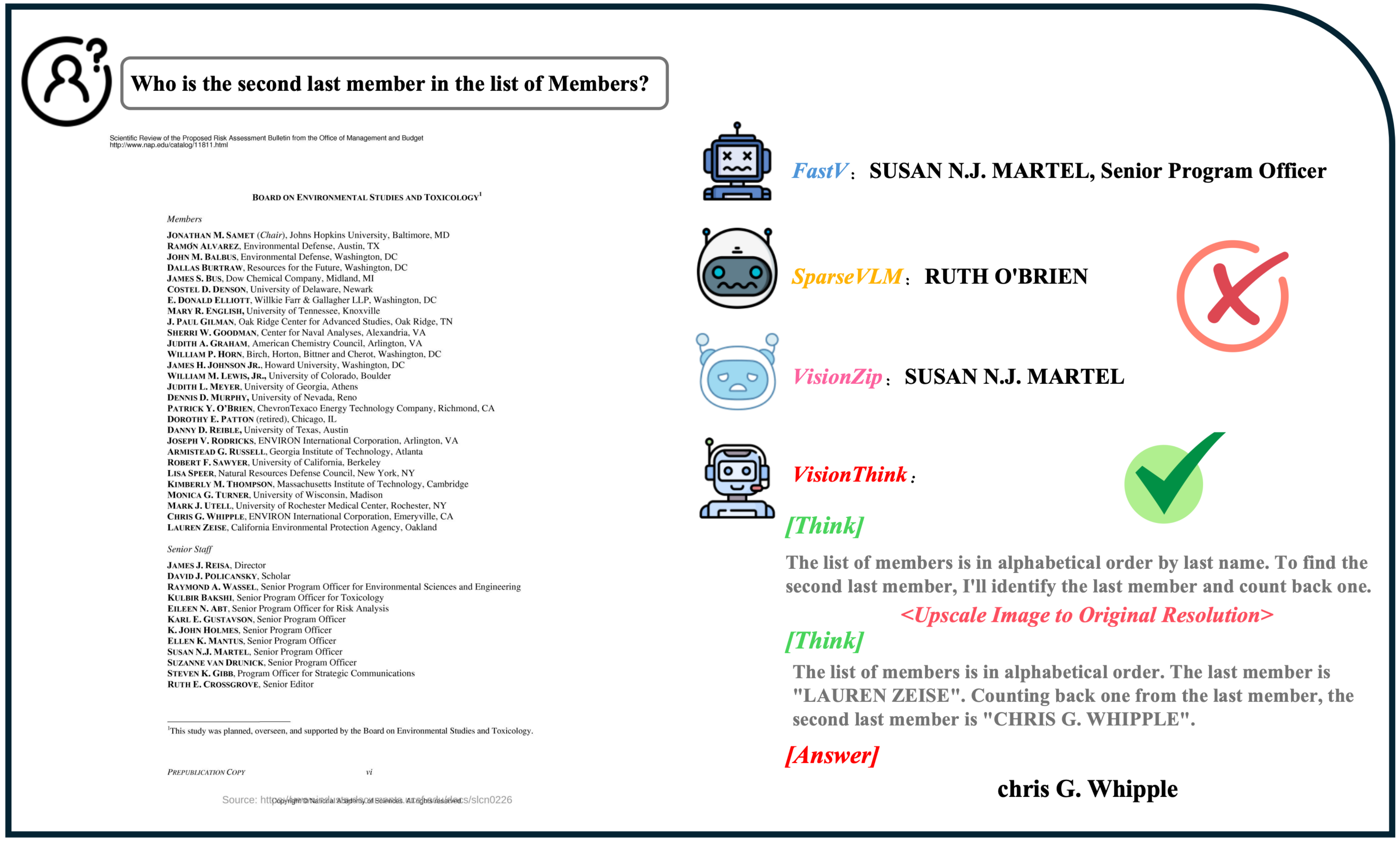}
    \label{fig:sup-case3}
\end{figure}

\begin{figure}[h]
    \centering
    \includegraphics[width=\linewidth]{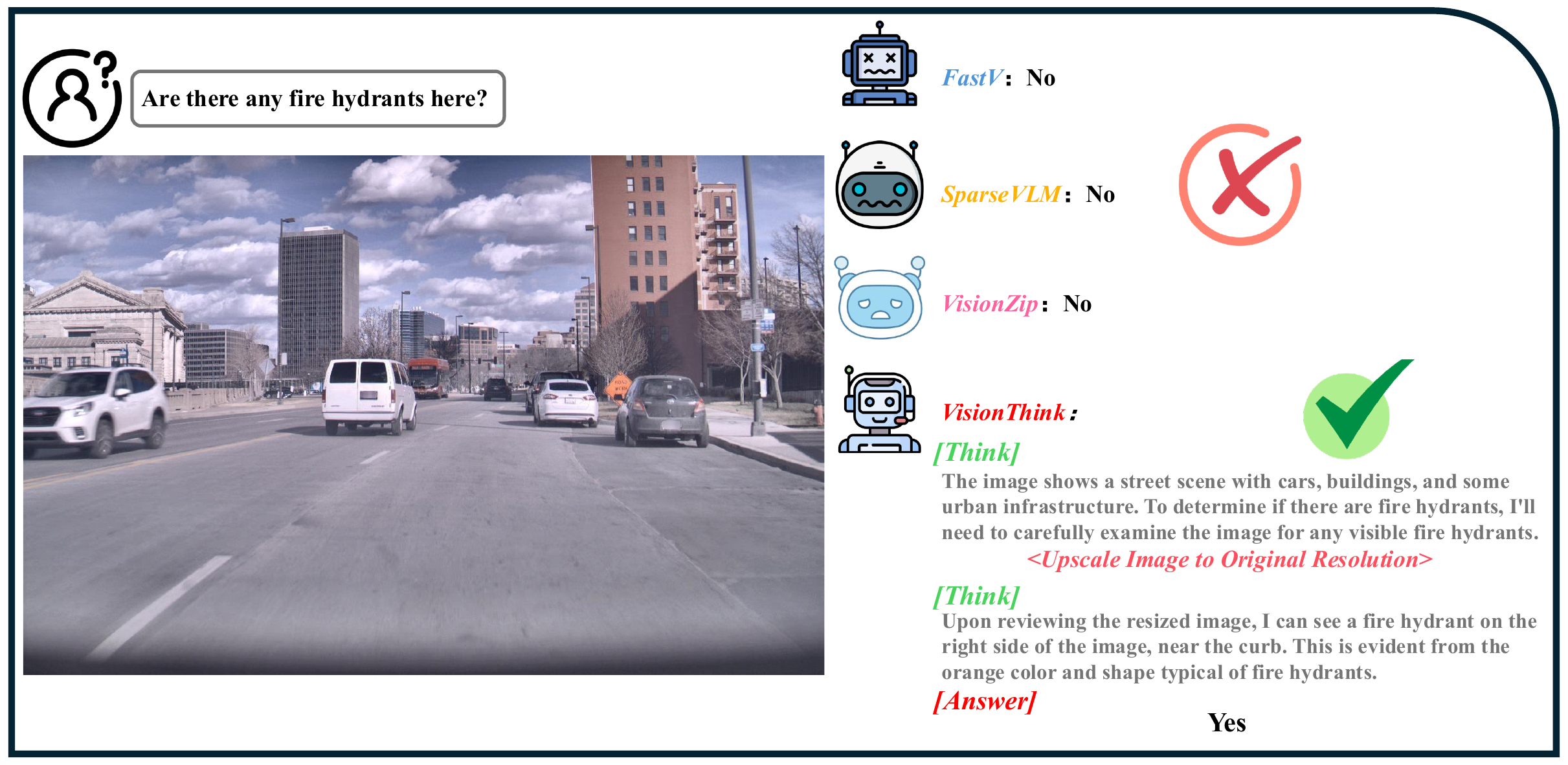}
    \label{fig:sup-case4}
\end{figure}

\section{Broader Impact Statement}
This paper is conducted solely for academic research purposes. All data used in this work were collected through compliant and ethical channels, ensuring adherence to relevant data protection and usage guidelines. Furthermore, all models employed in this study comply with their respective license agreements. As such, this research upholds high standards of integrity and responsibility, with no foreseeable negative societal impact.

\end{document}